\documentclass[a4paper,fleqn]{cas-sc}

\usepackage[authoryear,longnamesfirst]{natbib}
\usepackage{booktabs}
\usepackage{pifont}
\usepackage[flushleft]{threeparttable}
\usepackage{amsmath}

\newcommand{\accYes}{\ding{51}}
\newcommand{\cinterval}[2]{\,(#1--#2)}

\newcommand{\mq}[4][]{\shortstack{$#2^{#1}$\\[1pt]{\scriptsize$(#3,\,#4)$}}}
\newcommand{\nodata}{\shortstack{--\\[1pt]{\scriptsize\phantom{$(0.000,\,0.000)$}}}}

\begin{document}
\let\WriteBookmarks\relax
\def\floatpagepagefraction{1}
\def\textpagefraction{.001}
\shorttitle{A Latent Diffusion Framework for Volumetric Dose Prediction}
\shortauthors{J. Boyer-Chammard et~al.}

\title [mode = title]{Learning the Dose Manifold: A Latent Diffusion Framework for Volumetric Dose Prediction in Automated Radiotherapy Treatment Planning}

\author[1,2,3]{J. Boyer-Chammard}[orcid=0009-0004-9788-2122
]
\cormark[1]
\ead{jeanne.boyer-chammard@universite-paris-saclay.fr}

\credit{Conceptualization, Methodology, Software, Writing -- original draft}

\affiliation[1]{organization={Inserm U1355 RAMO-IT, Institut Gustave Roussy},
                addressline={114 Rue Edouard Vaillant}, 
                city={Villejuif},
                postcode={94800}, 
                state={Île-de-France},
                country={France}}
\affiliation[2]{organization={Department of Radiation Oncology, Centre Léon Bérard},
                addressline={28 rue Laënnec}, 
                city={Lyon},
                postcode={69008}, 
                state={Auvergne-Rhône-Alpes},
                country={France}}
\affiliation[3]{organization={R\&D Artificial Intelligence, Therapanacea},
                addressline={8-10 Avenue Ledru Rollin}, 
                city={Paris},
                postcode={75012}, 
                state={Île-de-France},
                country={France}}
\affiliation[4]{organization={Department of Radiation Oncology, Lausanne University Hospital},
                addressline={46 rue de Bugnon}, 
                city={Lausanne},
                postcode={1015}, 
                state={Vaud},
                country={Switzerland}}
\affiliation[5]{organization={Department of Medical Physics and Radiation Oncology, Mary Bird Perkins Cancer Center},
                addressline={4950 Essen Lane}, 
                city={Baton Rouge},
                postcode={70809}, 
                state={LA},
                country={USA}}
\affiliation[6]{organization={Computer Science Department, University of Crete},
                addressline={Voutes Campus}, 
                city={Heraklion},
                postcode={70013}, 
                state={Crete},
                country={Greece}}
\affiliation[7]{organization={Institute of Applied and Computational Mathematics, Foundation for Research and Technology – Hellas},
                addressline={Nikolaou Plastira 100, Vassilika Vouton}, 
                city={Heraklion},
                postcode={70013}, 
                state={Crete},
                country={Greece}}
\affiliation[8]{organization={Archimedes, Athena Research Center},
                addressline={1 Artemidos Street},
                city={Marousi},
                postcode={15125},
                state={Attica},
                country={Greece}}

\author[3]{R. Vauclin}
\credit{Methodology, Software}
\author[2]{M.C. Biston}
\credit{Data curation}
\author[1]{P. Maury}
\credit{Data curation, Validation, Investigation}
\author[4]{M. Gau}
\credit{Validation, Investigation}
\author[5]{S. Stathakis}
\credit{Validation, Investigation}
\author[2]{V. Grégoire}
\credit{Supervision}
\author[3]{N. Paragios}
\credit{Supervision, Project administration}
\author[1]{E. Deutsch}
\cormark[2]
\credit{Supervision}
\author[6,7,8]{N. Komodakis}
\cormark[2]
\credit{Supervision, Writing -- review \& editing}

\cortext[cor1]{Corresponding author}
\cortext[cor2]{Equally contributing senior authors}

\begin{abstract}
Automated treatment planning (ATP) aims to reduce the unwarranted inter-planner variability of manual planning, while improving treatment quality and efficiency. Most ATP pipelines rely on the prediction of the achievable three-dimensional dose distribution for a given anatomy and the subsequent optimization of machine parameters to reproduce it. Diffusion-based methods are largely slice-wise due to the high dimensionality of radiotherapy data, and assume a one-to-one mapping between anatomy and optimal dose despite the multi-objective nature of treatment planning. They are evaluated on their ability to match reference doses and rarely on their integration in ATP frameworks. We present LaDose, a latent diffusion framework for fully volumetric dose prediction. An autoencoder learned on dose distributions, trained so that perturbed latents decode to plausible doses, compresses dose maps 128-fold and enables 3D diffusion on a single GPU. Prescription values are represented as exact scalar vectors that bypass the image encoder, avoiding resampling and latent compression. Multiscale classifier-free guidance allows for the modulation of clinical trade-offs at inference, without retraining or additional labeled data. On 900 clinically delivered head-and-neck volumetric modulated arc therapy plans from five centers, LaDose reached a median absolute error of 1.18 Gy (1.4--1.8\% of the prescription) and was the only one of six frameworks free of systematic bias with respect to the reference distribution. Guidance scales acted predictably on target coverage and organ-at-risk sparing. Integrated with direct aperture optimization, the pipeline generated plans in 15.2 minutes; in a blinded review, senior clinicians rated 14 of 15 plans acceptable without editing.
\end{abstract}

\begin{graphicalabstract}
\includegraphics[width=1.0\textwidth]{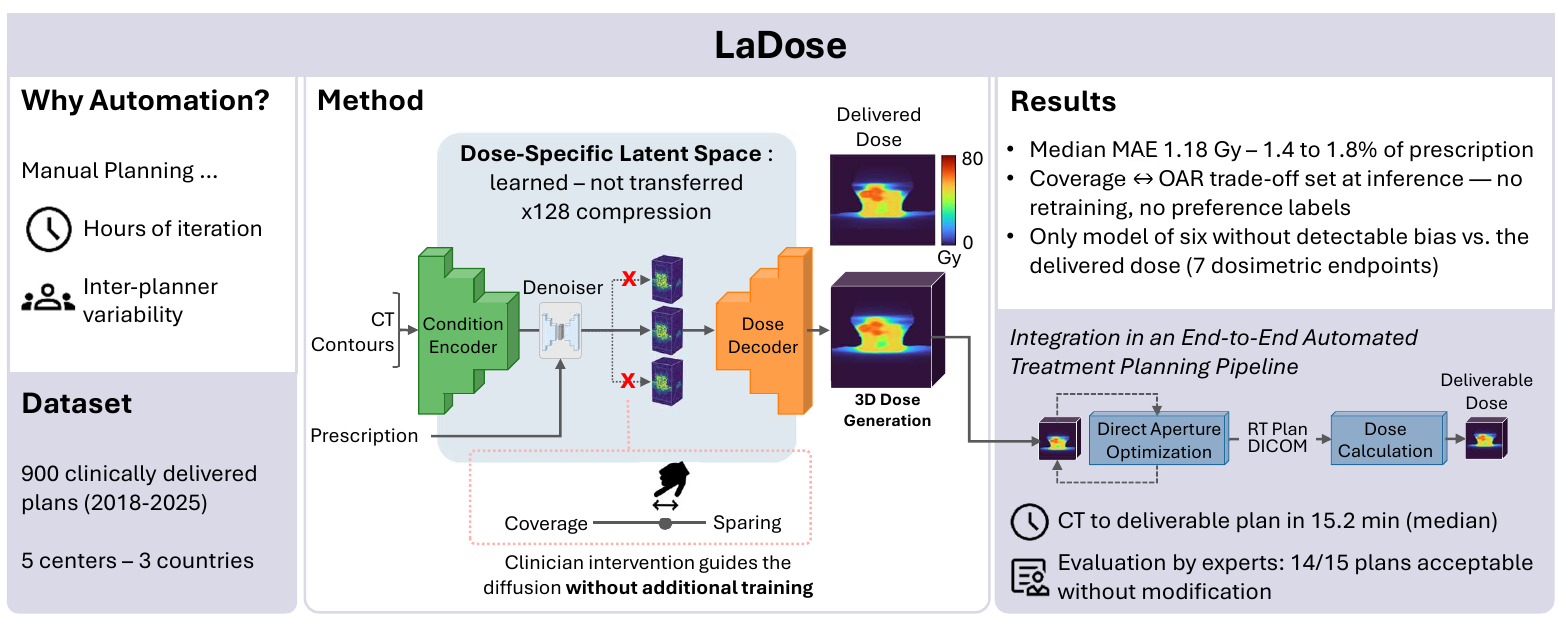}
\end{graphicalabstract}

\begin{highlights}

\item Learning a dose-specific latent space with high compression enables 3D diffusion.
\item The predictions of a single trained model are tunable by the user at inference.
\item Median error is 1.18 Gy on a dataset of 900 clinically approved treatment plans. 
\item Only model of six without detectable bias with respect to the delivered dose.
\item Plans built from the predicted dose were rated acceptable unedited in 14 of 15 cases.

\end{highlights}


\maketitle

\section{Introduction}

Radiotherapy is indicated in the management of an estimated 52.3\% of cancer patients \citep{delaney_role_2005}. Treatment planning is a highly complex and time-consuming inverse optimization process, aiming to deliver a prescribed dose to the planning target volume (PTV), while sparing surrounding organs at risk (OARs). Traditionally, dosimetrists manually adjust beam parameters and structure-specific objectives to iteratively refine the plan until it satisfies protocol constraints. As the trade-offs achievable for a given anatomy are not known a priori, this process is a manual search whose stopping point is determined by the time available and by the planner's experience \citep{husseinAutomationIntensityModulated2018}. As a consequence, manual planning introduces wide inter-planner variability, including in the compliance with the constraints that determine clinical acceptability \citep{nelmsVariationExternalBeam2012a}. Part of this variability reflects different institutional protocols and clinician preferences, which an automated replacement must take into account explicitly. The remainder stems from limited planning time and expertise, and is the variability that automated treatment planning (ATP) aims to remove through standardization, improving plan quality and consistency while reducing planning time \citep{mominKnowledgebasedRadiationTreatment2021b}.

Modern ATP pipelines often rely on direct aperture optimization with dose-mimicking objectives \citep{babierOpenKBPOptInternationalReproducible2022}. A predictor first estimates the achievable three-dimensional (3D) dose distribution for a given anatomy, and a downstream optimization engine then determines the machine parameters whose delivered dose best matches it. The rationale behind dose mimicking is that the predicted dose encodes the clinical trade-offs from previously approved treatment plans, so that the optimizer focuses on achievable, patient-specific targets rather than on generic constraints that a clinician would need to specify and tune manually \citep{dosemimicking}. Because the optimizer targets the predicted dose, errors in the prediction propagate into the delivered plan, making accurate prediction of the 3D dose distribution from the patient computed tomography (CT) scan, PTV volumes and OAR contours a critical intermediate step.

Learning-based dose prediction was initially formulated as a deterministic regression task from the planning CT scan and the delineated PTV and OARs, solved by leveraging convolutional encoder-decoder-based architectures such as the 3D U-Net \citep{C3D}. Because a given anatomy admits a set of clinically valid plans that differ in how trade-offs were prioritized, the optimum of a regression trained against a single plan is the conditional mean over that set, which blurs the prediction and smooths the steep gradients separating target volumes from healthy tissue \citep{fengDiffDPRadiotherapyDose2023}. To address this limitation, generative approaches were introduced, first with Generative Adversarial Networks (GANs) \citep{zhanMulticonstraintGenerativeAdversarial2022, gaoFlexibleCmGANPrecise2023}, and more recently with diffusion models \citep{fengDiffDPRadiotherapyDose2023, fengDiffusionbasedRadiotherapyDose2024, fuMDDoseDiffusionModel2025}. The poor scalability of diffusion models in voxel space limited their use to two-dimensional (2D) slices in which consistency across slices is not modeled explicitly. Concurrently with the present work, \citet{wang_any2any_2026} demonstrated tractable volumetric diffusion by operating in a latent space obtained by transfer from a frozen video autoencoder pretrained on natural images. Diffusion in a latent space learned on dose distributions themselves, rather than transferred, has to our knowledge not been investigated.

Whether deterministic or generative with fixed sampling, these methods return a single dose distribution for a given anatomy, and therefore commit to a specific set of choices in clinical trade-offs learned during training. In clinical practice, treatment planning requires balancing competing constraints with preferences and protocols varying across clinicians and institutions. The lack of efficient human-in-the-loop control has been identified as a challenge to the clinical integration of ATP \citep{gao_demo_2025}. Two recent frameworks make these trade-offs adjustable. \citet{gao_demo_2025} introduce a preference penalty within a GAN-based framework, and \citet{wang_any2any_2026} align predictions to institutional preferences through reinforcement learning post-training. In both cases, the preference specification must be integrated into the training procedure in a supervised way, and Any2Any further assumes fixed planning preferences per institution, which is not verified clinically \citep{nelmsVariationExternalBeam2012a}.

A second obstacle to the use of such models in clinics is how they are typically evaluated. Performance is almost always reported as the voxel-wise agreement with a single reference plan, together with dose-volume histogram (DVH)-derived error \citep{babierOpenKBPOpenaccessKnowledgebased2021a}. These quantities measure the reproduction of one acceptable plan among the many possibilities rather than the clinical acceptability of the prediction. They are computed on grids, crops and normalizations that differ between studies, making inter-study direct comparisons difficult \citep{kui_review_2024, zamanian_deep_2025}. Additionally, accuracy measured on the predicted dose does not guarantee the clinical acceptability of the plan generated from it. In their OpenKBP-Opt study, \citet{babierOpenKBPOptInternationalReproducible2022} report a moderate positive correlation (Spearman, $\rho$ between 0.5 and 0.62) between dose prediction performance and plan quality; a model outperforming another one on dose prediction metrics will not always yield a better plan. Conventional dose prediction evaluation metrics are therefore necessary but insufficient evidence of the clinical acceptability of the plan that is ultimately delivered, which is rarely evaluated or reported. 

In this work, we use a cohort of 900 clinically delivered volumetric modulated arc therapy (VMAT) plans coming from five centers in three countries, spanning a range of prescription levels, protocols and planning styles. The dataset retains the inter-institutional and inter-clinician variability which is removed by construction when using a dataset of plans generated by a single automated treatment planning solution as is the case in the public AAPM GDP-HMM cohort \citep{gaoAutomatingHighQuality2025}. Nonetheless, we also report results on this dataset to provide a benchmark against which future work can be compared directly.

We introduce \textbf{LaDose}, a radiotherapy-specific latent diffusion framework for fully 3D dose prediction, trained on 900 clinically delivered plans from five centers in three countries. In a six-framework comparison on this cohort, LaDose is the only model without a statistically significant bias across dosimetric endpoints, and the first latent diffusion predictor evaluated in a blinded expert review. It systematically addresses three core challenges: the representation in which generation occurs, the injection and control of clinical intent, and the clinical validity of the resulting plans. Our contributions are as follows:
\begin{enumerate} \item \textbf{A dose-specific latent space for whole-volume generation.}
We learn a latent space that captures the manifold of physically and clinically plausible dose maps. Because a dose distribution is a smooth field constrained by the physics of delivery, this manifold admits aggressive compression—reaching a factor of 128, compared with 16 for the dose latent space of concurrent work \citep{wang_any2any_2026}. Generation therefore proceeds on entire volumes on a single GPU, without spatial decomposition and without loss of global consistency. The framework overcomes the scalability limits of conventional diffusion approaches while embedding prior knowledge of valid dose distributions. This relieves the denoiser of the need to learn what a plausible dose distribution looks like, freeing its capacity for the conditioning-to-dose mapping.
\item \textbf{A generation-aware latent geometry.}
Latent diffusion pipelines typically optimize the autoencoder solely for reconstruction, leaving the geometry of the latent space unconstrained away from the encoder outputs—precisely the region occupied by sampled latents. We instead adapt latent denoising regularization \citep{yang2026latent} to dose distributions, exposing the autoencoder to the same denoising objective used by the downstream diffusion model. The resulting latent space keeps perturbed latents decodable to valid dose distributions, so that residual denoiser error manifests as small, physically plausible deviations in voxel space. This incurs a measurable cost in voxel-wise reconstruction fidelity (a quantity never queried at inference) and yields a substantial gain in the quality of the predicted doses.
\item \textbf{Disentangled conditioning.}
We separate target geometry from prescription information. Prescribed dose levels are supplied as exact scalar values rather than being painted into the voxel grid, where resampling, interpolation, and latent compression would otherwise degrade them.
\item \textbf{Multiscale classifier-free guidance.}
We introduce a guidance mechanism that modulates the influence of each conditioning signal at inference time. This enables prescription-level and planner-level control over the generated plan and accommodates inter-institutional planning diversity without per-institution retraining or additional supervised training.
\item \textbf{Evaluation on deliverable plans.}
We integrate the predictor with a dose-mimicking direct aperture optimization engine and report deliverable plan quality against the corresponding clinical plans for 30 patients treated from 2020 onwards, ensuring the comparison reflects current clinical practice. Acceptability is further assessed in a blinded review by three senior radiation oncologists and one senior medical physicist across three countries. 
\end{enumerate}

The remainder of this work is structured as follows. Section~\ref{sec:related} reviews related work on automated treatment planning, deep-learning-based dose prediction and generalization across institutions, especially in the head-and-neck (HN) region. Section~\ref{sec:methods} describes the proposed radiotherapy-specific latent representation, the associated 3D diffusion-based dose prediction model and its integration with a fully automated treatment planning pipeline. Section~\ref{sec:exp} details the dataset, implementation details, and the evaluation framework. Section~\ref{sec:results} reports reconstruction and prediction performance, along with the clinical acceptability of the generated plans. Section~\ref{sec:discussion} discusses the design choices, the scope of the clinical evaluation and the directions for future work.

\section{Related Work}\label{sec:related}

\subsection{Automated Treatment Planning and Dose Mimicking}
Knowledge-based planning (KBP) refers to the use of previously accepted and delivered plans to estimate an achievable dose distribution given a new anatomy. Initial KBP methods relied on geometric descriptors of the patient anatomy and PTV distribution \citep{wuPatientGeometrydrivenInformation2009} to infer organ-wise DVHs \citep{appenzollerPredictingDosevolumeHistograms2012}. The latter were then used as objectives for treatment planning systems (TPS). This approach brought improvements in terms of plan consistency and quality, and was adopted commercially \citep{hansenAutomaticTreatmentPlanning2016}. 

However, a DVH is a one-dimensional summary of a three-dimensional dose distribution which, by nature, does not encode spatial information; two identical DVHs can mask substantial differences in hot spot and gradient location. Because the prediction of the complete dose distribution removes this ambiguity, it has become the preferred formulation in current ATP pipelines \citep{babierOpenKBPOptInternationalReproducible2022}. Indeed, it provides an achievable voxel-wise objective for machine parameter optimization without requiring a clinician to specify them \citep{dosemimicking}. 

\subsection{From Regression to Generative Dose Prediction}
Most existing frameworks formulate dose prediction as a deterministic voxel-wise regression task from CT; such approaches predominantly leverage U-Net-based architectures and their variants \citep{kontaxisDeepDoseFastDose2020, Deeplab, C3D}, e.g., augmented with transformer components \citep{huTrDosePredDeepLearning2023}. Fully transformer-based architectures have been explored \citep{wenTransformerEmbeddedMultiTaskModel2023a} but are often limited to two-dimensional settings due to their high computational cost, a limitation that can be alleviated through latent-space compression of the input data \citep{wangLatentSpacesEnable2024}. Despite their success, regression-based methods are known to suffer from high-frequency detail loss \citep{fengDiffDPRadiotherapyDose2023} and fail to integrate prior knowledge on the clinical and physical properties of valid dose distributions \citep{feng_leveraging_2026}. This has led recent studies to formulate dose prediction as a generative modeling task. Approaches based on GANs \citep{zhanMulticonstraintGenerativeAdversarial2022, gaoFlexibleCmGANPrecise2023} demonstrated improved local detail reconstruction, but are prone to unstable training and mode collapse \citep{dhariwalDiffusionModelsBeat2021}. Diffusion models (DMs) have shown superior performance in recovering high frequency details, leading to their adoption in dose prediction pipelines \citep{fengDiffDPRadiotherapyDose2023, fengDiffusionbasedRadiotherapyDose2024, fuMDDoseDiffusionModel2025}. However, the computational cost of diffusion models in voxel space has confined most work to a slice-wise formulation. Concurrently with the present work, \citet{wang_any2any_2026} achieved fully volumetric diffusion-based dose prediction by operating in a latent space with a compression rate of 16. They leverage a frozen video autoencoder pretrained on natural images (Wan~2.1, 127~M parameters).

\subsection{Conditioning Signals}

The most common setting conditions on the planning CT, together with the PTV and OAR contours \citep{C3D,nguyen3DRadiotherapyDose2019, gronbergTechnicalNoteDose2021}. Other models also supply geometric descriptors derived from the masks, in the form of signed distance maps, based on the hypothesis that the distance to the target boundary is a strong predictor of the dose distribution \citep{dosediff,cao_beam_2025}. A third family of approaches leverages the geometry of the delivered plan. \citet{teng_beam-wise_2024} generate binary beam masks to allow for downstream beam-wise decomposition and refinement of the predicted dose. The publicly available GDP-HMM dataset \citep{gaoAutomatingHighQuality2025} provides beam plates and an additional input modality, encoding delivery geometry \citep{xiongGeneralizableDosePrediction2026,wang_any2any_2026}. \citet{cao_beam_2025} combine beam trajectories with pencil-beam dose calculation to ground the prediction in the physics of the delivery. Reconstruction performance improves with this conditioning strength progression, along with the preprocessing burden. 

The handling of the prescription condition is largely inherited from the dataset on which each method is developed. The OpenKBP dataset \citep{babierOpenKBPOpenaccessKnowledgebased2021a} contains a single prescription protocol (70/63/56~Gy). Associated released implementations commonly encode prescription by assigning the prescribed dose to the associated PTV contour \citep{C3D, dosediff}. Uniform prescription protocols render the prescription condition uninformative and absorb it into the constant dose normalization \citep{fuMDDoseDiffusionModel2025}. The GDP-HMM dataset \citep{gaoAutomatingHighQuality2025} contains multiple prescription protocols, and the prescription is encoded by assigning the prescribed dose to the associated PTV contour \citep{wang_any2any_2026}. When supplied, the prescribed dose is carried by a spatial input and undergoes the same resampling, interpolation and compression in the case of a latent formulation \citep{wang_any2any_2026}.

\subsection{Clinical Preference Integration} Two recent frameworks allow for the integration of clinical preferences. \citet{gao_demo_2025} introduce a homogeneity index and per OAR sliders, which can be manually tuned at inference time by the user to adjust OAR sparing or target coverage. To do so, they sample slider values during training and add a secondary supervised objective in the loss function of the model, which correlates slider values to dosimetric endpoints. \citet{wang_any2any_2026} align their dose prediction model to specific institutes via an additional reinforcement learning step, which requires retraining the model for each new institution.

\subsection{Generalization Across Institutions and Protocols}
Historically, dose prediction frameworks have been trained and evaluated on single-center cohorts of anatomical sites such as lung or prostate, or on intensity-modulated radiation therapy (IMRT) delivery. In an effort to enable generalization, the GDP-HMM challenge released a large dataset of IMRT and VMAT plans for multiple prescription protocols across the head-and-neck and thoracic regions \citep{gaoAutomatingHighQuality2025}. The plans are, however, generated automatically and thus follow a single, consistent planning style. The authors report that they have not been clinically approved at this stage. 

Due to the high density of critical anatomical structures and the use of VMAT as the predominant delivery mode \citep{alfouzanRadiationTherapyHead2021}, accurate dose delivery is particularly challenging in the HN region \citep{zamanian_deep_2025}. Moreover, heterogeneous evaluation practices and limited data availability further complicate progress in HN dose prediction \citep{kui_review_2024}.

\section{Methods}\label{sec:methods}

\subsection{Preliminaries}

\paragraph{Diffusion Models}
Diffusion models generate samples by learning to reverse a stochastic corruption process. Let \(x_0 \in \mathbb{R}^{D \times H \times W}\) denote a clean sample from the target data distribution. Following the formulation of \citet{karrasElucidatingDesignSpace2022}, a noisy sample at noise level \(\sigma\) is given by
\begin{equation}
x_\sigma = x_0 + \sigma \varepsilon,
\qquad
\varepsilon \sim \mathcal{N}(\mathbf{0},\mathbf{I}),
\label{eq:diffusion_forward}
\end{equation}

where during training, \(\sigma\) is sampled from the log-normal distribution. A denoising network \(\mathcal{D}_{\theta}\) predicts the clean sample according to

\begin{equation}
\hat{x}_0 = \mathcal{D}_{\theta}(x_\sigma,\sigma,\mathcal{C}),
\label{eq:denoiser}
\end{equation}

where \(\mathcal{C}\) denotes optional conditioning information. For deterministic sampling, the reverse process can be expressed as the probability-flow ordinary differential equation (ODE)
\begin{equation}
\frac{\mathrm{d}x}{\mathrm{d}\sigma}
=
\frac{x - \mathcal{D}_{\theta}(x,\sigma,\mathcal{C})}{\sigma}.
\label{eq:pf_ode}
\end{equation}

\paragraph{Classifier-Free Guidance}
Classifier-free guidance (CFG) enables conditional generation by jointly training the denoiser with conditional and unconditional inputs \citep{ho2021classifierfree}. At inference, the conditional prediction is amplified relative to the unconditional prediction as
\begin{equation}
\mathcal{D}_{\theta}^{\mathrm{CFG}}(x_\sigma,\sigma,\mathcal{C})
=
\mathcal{D}_{\theta}(x_\sigma,\sigma,\varnothing)
+
s
\left[
\mathcal{D}_{\theta}(x_\sigma,\sigma,\mathcal{C})
-
\mathcal{D}_{\theta}(x_\sigma,\sigma,\varnothing)
\right]
\label{eq:cfg}
\end{equation}
where \(s \geq 0\) is the guidance scale controlling the strength of conditioning.

\paragraph{VMAT Delivery and Treatment Planning}
External beam radiotherapy delivers ionizing radiation to the patient using a linear accelerator (LINAC). The LINAC gantry rotates around the patient, while the photon beam is shaped by a multi-leaf collimator (MLC) consisting of opposing rows of motorized tungsten leaves. The delivered radiation is quantified in monitor units (MU). In VMAT, gantry rotation, MLC leaf positions, and dose rate vary continuously and simultaneously \citep{alfouzanRadiationTherapyHead2021}. For treatment planning, delivery is discretized into control points, each defining the gantry angle, dose rate, and MLC aperture. Inverse planning determines a sequence of control points whose accumulated dose achieves adequate target coverage while satisfying the clinical constraints imposed by neighboring OARs.

\subsection{Radiotherapy-Specific 3D Latent Representation}\label{sec:latent}

Diffusion directly in voxel space is computationally prohibitive at the resolution of clinical dose distributions and therefore requires a compact latent representation \citep{rombachHighResolutionImageSynthesis2022}. The choice of latent space determines both the computational cost of diffusion and the complexity of the distribution that the denoising network must model. We therefore learn a radiotherapy-specific latent space directly from dose distributions rather than adopting a general-purpose latent representation designed for natural images.

We use variational autoencoders (VAEs) to encode the different input modalities into compact volumetric representations. The diffusion process is performed entirely in the latent space, while a dedicated dose decoder maps generated dose latents back to voxel space. Because dose distributions are smooth scalar fields governed by the physics of radiation delivery, their intrinsic dimensionality is substantially lower than that of natural images. A dose-specific latent space can therefore exploit this structure to achieve high compression while retaining the spatial detail required for accurate dose prediction. In addition, the dose decoder implicitly captures the prior of plausible dose distributions, allowing the diffusion model to focus primarily on learning the conditional mapping from patient and treatment information to dose.

\paragraph{Variational Autoencoding of Volumetric Representations}
Each input modality is encoded using a dedicated VAE with encoder \(\mathcal{E}\) and decoder \(\mathcal{G}\),
\begin{equation}
z = \mathcal{E}(x),
\qquad
\hat{x} = \mathcal{G}(z),
\label{eq:vae}
\end{equation}
where \(z \in \mathbb{R}^{c \times d \times h \times w}\) is a volumetric latent representation. The diffusion process is defined entirely in the latent space, and the generated dose latent is subsequently mapped to voxel space using the dose decoder \(\mathcal{G}_{\mathrm{dose}}\).

All VAEs are trained using a reconstruction objective with Kullback--Leibler (KL) regularization. Specifically, the base VAE objective is
\begin{equation}
\begin{aligned}
\mathcal{L}_{\mathrm{VAE}}
&=
\mathbb{E}_{x \sim p_{\mathrm{data}}}
\left[
\frac{\mathcal{L}_{\mathrm{rec}}(x,\hat{x})}{\exp(\log \sigma_{\mathrm{rec}}^{2})}
+
\log \sigma_{\mathrm{rec}}^{2}
\right]
\\
&\quad +
\beta \,
\mathrm{KL}
\left(
q_{\phi}(z \mid x)
\,\middle\|\,
\mathcal{N}(\mathbf{0},\mathbf{I})
\right),
\end{aligned}
\label{eq:vae_loss}
\end{equation}
where \(\sigma_{\mathrm{rec}}\) is a learned, network-specific reconstruction scale and \(\beta\) controls the strength of the KL regularization. The base reconstruction loss \(\mathcal{L}_{\mathrm{rec}}\) is an \(L_1\) loss, augmented with modality-specific terms.

For dose distributions, which contain steep spatial gradients near target boundaries, we additionally use a gradient-consistency loss
\begin{equation}
\mathcal{L}_{\nabla}(x,\hat{x})
=
\left|
\nabla x - \nabla\hat{x}
\right|_1,
\label{eq:gradient_loss}
\end{equation}
following \citet{gradloss}. The dose VAE is trained at a compression factor of 128. CT volumes are encoded at a compression factor of 64 and use both the gradient-consistency loss and a perceptual loss,
\begin{equation}
\mathcal{L}_{\mathrm{LPIPS}}(x,\hat{x})
=
\lambda_{\mathrm{LPIPS}} \,
\mathrm{LPIPS}(x,\hat{x}).
\label{eq:lpips_loss}
\end{equation}
The CT VAE additionally employs an adversarial loss,
\begin{equation}
\mathcal{L}_{\mathrm{adv}}
=
\lambda_{\mathrm{GAN}} \,
w_{\mathrm{adaptive}} \,
\mathcal{L}_{\mathrm{GAN}}(\hat{x}),
\label{eq:adv_loss}
\end{equation}
which is applied outside the reconstruction objective and combined with it using a gradient-based adaptive weight $w_{\mathrm{adaptive}}$. Binary target masks are encoded at a compression factor of 256 using a foreground-weighted \(L_1\) loss to preserve small structures. All encoders use the same spatial downsampling factor, allowing their latent representations to be concatenated channel-wise despite their different overall compression factors.

\paragraph{Generation-Aware Latent Space Regularization}\label{sec:denoreg}
In latent diffusion models, the autoencoder is typically optimized independently of the downstream generative model. This creates a potential mismatch between training and inference: the decoder is trained primarily on $z_{0}$, the latent representations produced by the encoder, whereas during diffusion sampling it receives \(\hat{z}_0\), an estimate of the clean latent obtained after a finite number of denoising steps. Consequently, \(\hat{z}_0\) may lie near, but not exactly on, the encoder’s latent manifold. If the decoder is poorly behaved in this neighborhood, small denoising errors can result in disproportionately large reconstruction errors in voxel space.

We address this issue by explicitly shaping the latent space so that neighborhoods around encoded samples remain meaningful and smoothly decodable. Inspired by \citet{yang2026latent}, training of the dose visual tokenizer is performed in two stages. In the first stage, encoder latents are interpolated with samples from a Gaussian prior before being passed to the decoder. For an encoded latent \(z\), we define
\begin{equation}
z'
=
(1-\tau)z
+
\tau\gamma\varepsilon,
\qquad
\varepsilon\sim\mathcal{N}(\mathbf{0},\mathbf{I}),
\qquad
\tau\sim\mathcal{U}(0,1),
\label{eq:latent_corruption}
\end{equation}
where \(\gamma\) controls the standard deviation of the Gaussian component and \(\tau\) determines the corruption level. The decoder is trained to reconstruct the original input from \(z'\).

This latent corruption objective serves two complementary purposes. First, it regularizes the latent representation by discouraging information from being encoded in directions that are highly sensitive to small perturbations, resulting in a smoother latent geometry. Second, it aligns the autoencoder with the downstream diffusion task: the decoder is explicitly trained to map perturbed latent representations to valid dose distributions, rather than being optimized solely for reconstruction of encoder outputs. Consequently, deviations in latent space are more likely to produce gradual and physically plausible changes in the reconstructed dose.

The first training stage exposes the decoder to the full range of corruption levels through the uniform sampling of \(\tau\). However, inference primarily requires robustness to relatively small deviations around the clean latent, since the decoder receives the denoiser’s estimate \(\hat{z}_0\) rather than an arbitrarily corrupted latent. Training exclusively on corrupted latents can therefore trade reconstruction fidelity for robustness in regions of latent space that are less relevant at inference. To recover high-fidelity reconstruction, we use a second training stage in which the encoder is frozen and the decoder is fine-tuned using clean latents only. The decoder is thus initialized from a solution that is robust to latent perturbations, while the second stage restores reconstruction fidelity on the encoder manifold.

\subsection{3D Latent Diffusion for Dose Prediction}

\paragraph{Conditional 3D Latent Diffusion}
Dose prediction is formulated as conditional generation in the dose-specific latent space described in Section~\ref{sec:latent}. Let \(z_0\) denote the clean latent representation of the ground-truth dose distribution. The denoising network \(\mathcal{D}_{\theta}\) is trained to recover \(z_0\) from its noisy counterpart, conditioned on the latent representations of the planning CT, target structures, OARs, and prescription information.

\begin{figure*}
	\centering
	\includegraphics[width=\textwidth]{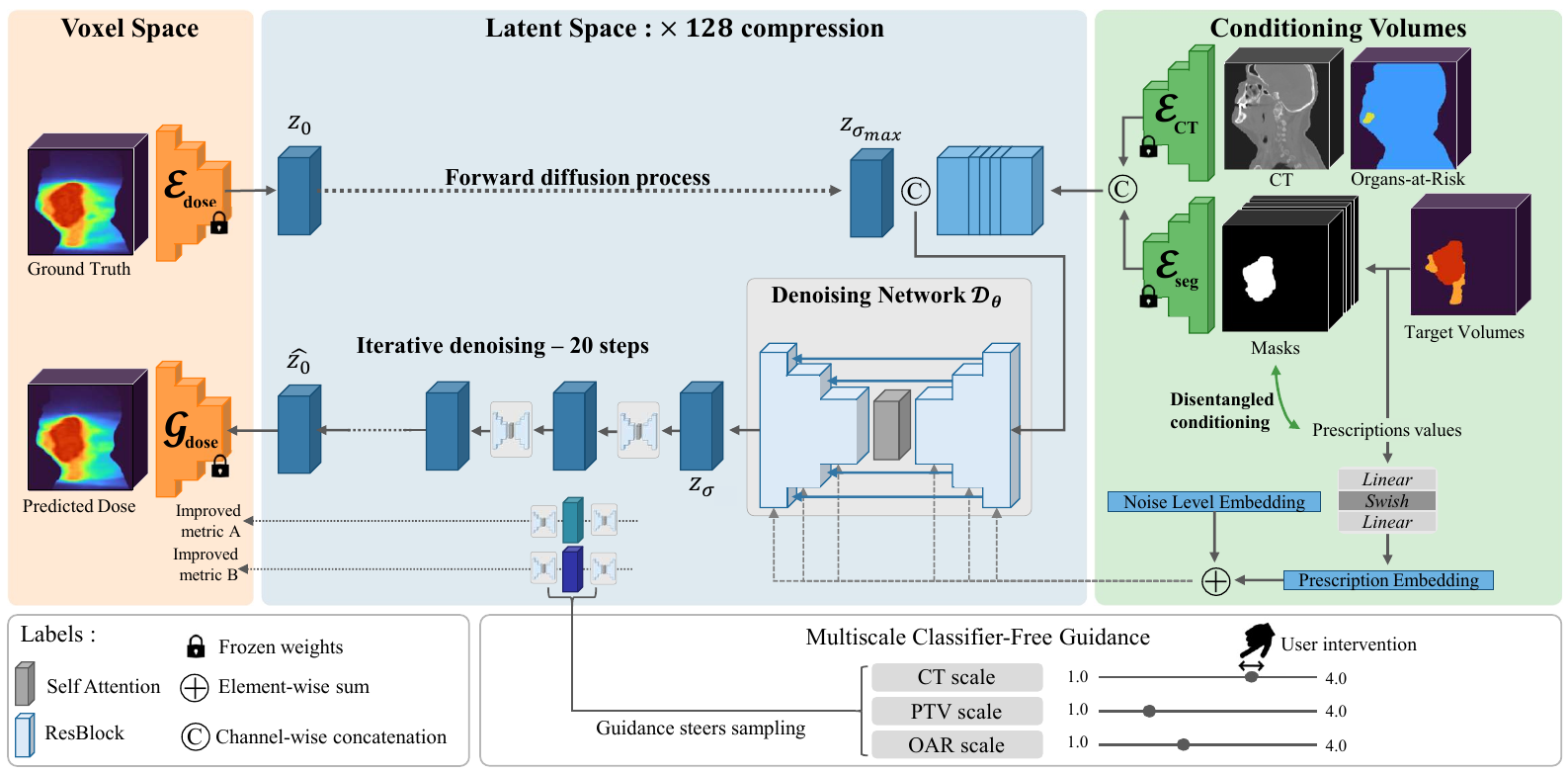}
	\caption{Overview of the proposed 3D latent diffusion framework for dose prediction. Patient and treatment information are encoded into volumetric latent representations that condition a 3D diffusion model. The generated dose latent is subsequently decoded into a voxel-space dose distribution.}
	\label{fig:overview}
\end{figure*}

The denoising network $\mathcal{D}_{\theta}$ is a three-dimensional U-Net with three resolution levels. Each level contains three residual blocks, with a self-attention layer at the bottleneck. The base channel width is 192, with channel multipliers of \(1\), \(2\), and \(2\) across the three resolutions. Training uses the \(\mathcal{L}_2\) denoising objective together with the preconditioning scheme and log-normal noise distribution of \citet{karrasElucidatingDesignSpace2022}. During inference, the generated dose latent is mapped to voxel space using the dose decoder. An overview of the complete framework is shown in Fig.~\ref{fig:overview}.

\paragraph{Disentangled Spatial and Prescription Conditioning}
We disentangle spatial and scalar conditioning to preserve prescription information while enabling spatial conditioning to propagate throughout the diffusion network. Jointly encoding a PTV volume with its associated prescription value would subject the scalar prescription information to spatial compression and may therefore degrade its numerical fidelity. Instead, spatial and scalar information are represented separately.

Spatial conditioning is introduced by concatenating the latent embeddings of the PTV and other spatial structures with the noisy dose latent. Prescription information is represented independently as a vector \(p\in\mathbb{R}^{3}\) and transformed by a two-layer multilayer perceptron (MLP) into a conditioning embedding \(e_p\). Let \(e_t\) denote the timestep embedding. The two embeddings are combined as
\begin{equation}
e = \mathrm{SiLU}(e_t + e_p),
\label{eq:conditioning_embedding}
\end{equation}
and projected into the feature space of each residual block. For an intermediate feature map \(h\), the resulting conditioning is injected additively as
\begin{equation}
h'= h + W_e e,
\label{eq:conditioning_injection}
\end{equation}
where \(W_e\) is a learned projection. Applying this modulation throughout the network ensures that prescription information is available at all depths while preserving its numerical representation independently of spatial compression.

\paragraph{Hierarchical Classifier-Free Guidance for Dose Control}
Dose prediction involves multiple conditioning signals with different clinical roles, and the desired balance between target coverage and OAR sparing may vary across patients and clinicians. To enable user-specific control over the predicted dose distribution, we implement multiscale classifier-free guidance \citep{brooksInstructPix2PixLearningFollow2023}, which allows for the independent modulation of each conditioning modality at inference time.

Let \(z_\sigma\) denote a noisy dose latent at noise level \(\sigma\), and let
\begin{equation}
\hat{z}_0^{\mathcal{C}} = \mathcal{D}_{\theta}(z_\sigma,\sigma,\mathcal{C})
\label{eq:conditional_prediction}
\end{equation}
denote the predicted clean dose latent under conditioning set \(\mathcal{C}\). We define the conditioning hierarchy
\begin{equation}
\varnothing \rightarrow \mathrm{CT} \rightarrow (\mathrm{CT},\mathrm{PTV}) \rightarrow (\mathrm{CT},\mathrm{PTV},\mathrm{OAR}),
\label{eq:conditioning_hierarchy}
\end{equation}
and compute the guided prediction as
\begin{equation}
\begin{aligned}
\tilde{z}_0 &= \hat{z}_0^{\varnothing} + s_{\mathrm{CT}} \left( \hat{z}_0^{\mathrm{CT}} - \hat{z}_0^{\varnothing} \right) \\
&\quad + s_{\mathrm{PTV}} \left( \hat{z}_0^{\mathrm{CT},\mathrm{PTV}} - \hat{z}_0^{\mathrm{CT}} \right) \\
&\quad + s_{\mathrm{OAR}} \left( \hat{z}_0^{\mathrm{CT},\mathrm{PTV},\mathrm{OAR}} - \hat{z}_0^{\mathrm{CT},\mathrm{PTV}} \right),
\end{aligned}
\label{eq:multiscale_cfg}
\end{equation}
where \(s_{\mathrm{CT}},s_{\mathrm{PTV}},s_{\mathrm{OAR}}\geq 0\) are independently adjustable guidance scales. Each scale controls the incremental contribution of the corresponding conditioning signal relative to the preceding level in the hierarchy. This formulation enables personalized dose generation without retraining the diffusion model or requiring additional labeled data.

\subsection{Direct Aperture Optimization with Dose Mimicking Objectives}\label{sec:dao}

\begin{figure*}
	\centering
	\includegraphics[width=0.8\textwidth]{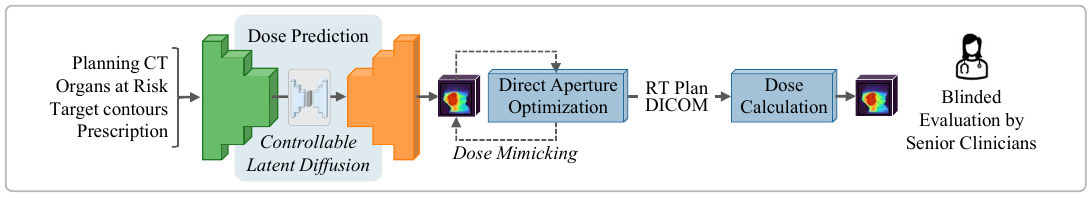}
	\caption{End-to-end automated treatment planning framework.}
	\label{fig:endtoend}
\end{figure*}

Plans are generated with a commercially available, fully automated VMAT planning
engine \citep{prunaretty2024}, based on direct aperture optimization (DAO) (Fig.~\ref{fig:endtoend}), which optimizes machine parameters directly rather than intermediate fluence maps. Let $\hat{d}\in\mathbb{R}^{N}_{+}$ denote the generated dose resampled to the planning CT grid. Let $x=(\ell,\mu)$ be the concatenated MLC leaf positions $\ell$ and the MU increments $\mu$ of all control points of the arcs. With $d(x)$ denoting the dose delivered by $x$, the machine parameters are obtained by solving the following optimization problem:

\begin{equation}
\min_{x \in \mathcal{X}} \;
\underbrace{\sum_{s\in\mathcal{O}} \frac{w_{s}}{|s|}\sum_{i\in s}
\Big[\, d_{i}(x)-\hat{d}_{i} \,\Big]_{+}^{2}}_{\text{dose mimicking}}
\;+\;
\underbrace{\sum_{s\in\mathcal{T}} \frac{w_{s}}{|s|}\sum_{i\in s}
\big( d_{i}(x)-p_{s} \big)^{2}}_{\text{target coverage}}
\;+\; \mathcal{R}(x),
\label{eq:dao}
\end{equation}

where $[\,\cdot\,]_{+}=\max(\cdot,0)$, $\mathcal{O}$ is the set of OAR contours and auxiliary structures such as overlaps, $\mathcal{T}$ the set of targets, $p_{s}$ the prescribed dose of target $s$ and $w_{s}$ a structure weight. The predicted dose acts as voxel-wise upper bound on healthy tissue: the model is penalized when it exceeds the sparing that the dose prediction model considers achievable for this anatomy. Target volumes are optimized independently from the prediction to ensure that coverage is dependent on protocol rather than implicitly passed as a voxel-wise objective. For the left and right brachial plexuses, along with the spinal cord, an additional fixed clinical constraint is added as a one-sided quadratic penalty at fixed dose levels.

To ensure the deliverability of the generated plans, the optimization is constrained to the set of machine-deliverable parameters $\mathcal{X}$ and regularized by $\mathcal{R}(x)$. $\mathcal{R}(x)$ comprises a penalty term on total MUs, a total-variation term smoothing the MU profile along each arc, and a leaf-edge term penalizing irregular aperture shapes. The set of machine-deliverable parameters $\mathcal{X}$ is defined by the minimum and maximum MLC leaf gap, the leaf-carriage span, and the maximum leaf travel and MU per segment derived from the leaf-speed, gantry-speed and dose-rate limits of the treatment unit. The dose is calculated with a continuous aperture model that integrates the fluence delivered between consecutive control points rather than sampling static apertures at discrete gantry angles \citep{dao_2018}, from dose-influence matrices precomputed with a collapsed-cone convolution dose engine \citep{ahnesjo1989}. The optimization problem is initialized with conformal apertures obtained from the beam's-eye-view projection of the targets, refined by column generation \citep{Peng_columngen_2012} and solved using first-order gradient descent. The final dose is recomputed with the same collapsed-cone dose engine at full resolution.

\section{Experimental Setup}\label{sec:exp}

\subsection{Dataset}\label{sec:dataset}

This study is based on a multi-centric, retrospective dataset of 900 head-and-neck cancer 
patients treated with VMAT between 2018 and 2025, from five centers across three countries. Prescriptions ranged from 50 to 70~Gy across one to three levels of PTV. The cohort was split into training (n=753), validation (n=71), and test (n=76) sets. For each patient, data included planning CT, clinically delivered dose distribution, and tumor and OAR contours (brainstem, spinal cord, parotids, mandible). The distribution of primary tumor sites and of prescribed dose levels is reported in Table~\ref{tab:cohort}. To our knowledge, this is the largest reported cohort of clinically approved VMAT plans used for dose prediction. 

Model performance was further assessed on the public AAPM GDP-HMM dataset \citep{gaoAutomatingHighQuality2025} restricted to the HN VMAT patients (472 training, 74 validation), with prescriptions from 50 to 70~Gy over one to three PTV levels. This cohort differs from ours in terms of planning protocol: plans were automatically generated by a single planning framework, and are correspondingly more homogeneous (median homogeneity index $4.27$ against $7.67$ percentage points on our in-house data).

\begin{table}[t]
\caption{Composition of the in-house cohort ($n=900$): distribution of tumor
primary sites (left) and of prescribed dose levels (right). Proportions are
percentages of the full cohort and sum to 100\% in each panel. Prescription
groups refer to the dose delivered to the high-dose PTV.}
\label{tab:cohort}
\centering
\begin{minipage}[t]{0.48\linewidth}
\centering
\begin{tabular}{l r}
\toprule
\textbf{Tumor primary site} & \textbf{Proportion (\%)} \\
\midrule
Oropharynx         & 40 \\
Hypopharynx/Larynx & 30 \\
Oral cavity        & 25 \\
Others             & 5 \\
\bottomrule
\end{tabular}
\end{minipage}\hfill
\begin{minipage}[t]{0.48\linewidth}
\centering
\begin{tabular}{l r}
\toprule
\textbf{Prescription (Gy)} & \textbf{Proportion (\%)} \\
\midrule
$\leq 55$    & 21.1 \\
$\approx 60$ & 13.6 \\
$\approx 65$ & 21.7 \\
$\approx 70$ & 43.6 \\
\bottomrule
\end{tabular}
\end{minipage}
\end{table}

\subsection{Evaluation Metrics and Protocol}\label{sec:eval}

All models assessed on our in-house dataset were trained at a single resolution and evaluated under a single protocol. No resampling was performed at inference time.
Performance was assessed using the mean absolute error (MAE) and the DVH metric MAE from the OpenKBP challenge \citep{babierOpenKBPOpenaccessKnowledgebased2021a}. As these metrics do not reflect the asymmetrical clinical impact of errors, we also report the Homogeneity Index (HI, HI $=(D_{2}-D_{98})/D_{50}\times100$), the Nakamura Conformity Index (nCI) \citep{nakamura_dose_2001} and $D_{99}$, or dose received by 99\% of the high-dose PTV to assess tumor coverage, as well as standard OAR DVH metrics to evaluate OAR sparing. The metrics were computed on the set of voxels restricted to the patient external contour, excluding the couch and immobilization devices. As the objective of the dose prediction step is to accurately reconstruct the clinical dose, we report the signed deviation of the predicted from the delivered dose of these additional metrics, rather than the absolute error. Evaluation on the AAPM GDP-HMM dataset was carried out using the same protocol, but in the geometry of the AAPM GDP-HMM dataset so that it is directly comparable with future work. 

The unit of analysis is the patient. Statistical significance was assessed using two-sided Wilcoxon signed-rank tests, between each baseline and LaDose, as well as between each model and zero to assess systematic bias. $p$-values are adjusted using the Holm--Bonferroni procedure.

\subsection{Baseline Models}

We compare against four published dose prediction models spanning the two dominant formulations: two convolutional regression networks, C3D \citep{C3D} and DCNN \citep{dcnn}, and two diffusion-based models, DoseDiff \citep{dosediff} and MD-Dose \citep{fuMDDoseDiffusionModel2025}. Additionally, we include the winning entry of the AAPM 2025 GDP-HMM dose prediction challenge \citep{gdpwinner}, based on MedNeXt \citep{mednext}, as it defines the current state of the art on the largest public dose prediction dataset \citep{gaoAutomatingHighQuality2025}. We reproduce the single MedNeXt model rather than the challenge-winning ensemble of that model with the organizers' baseline network. Each model was retrained from scratch on the training split of our cohort using the authors' released implementation, and the best-performing checkpoint was selected for evaluation. Hyperparameters were kept as reported by the authors unless the original configuration was incompatible with our data or the model did not train properly. The adaptations fall into two main categories, and are reported in detail per model in Appendix~\ref{app:baselines}.

First, the field of view and model resolution was adapted to match that of our dataset. Second, prescription conditioning was introduced in implementations that assumed a uniform protocol. Additional changes were made when the released code contained defects that prevented training or inference: for instance, the released MD-Dose implementation sampled the latent prior using min-max normalization, which introduced a mismatch with the noise injected during training and resulted in a high MAE on our test set. Any correction improved the baseline, and we report the best achieved results throughout.

On the AAPM GDP-HMM dataset, we compare against the winner's released checkpoint on the challenge's own data, including the beam geometry conditioning \citep{gaoAutomatingHighQuality2025,gdpwinner}. The two models operate under different conditioning regimes which favor the challenge model. 

\subsection{LaDose Implementation}

\subsubsection{Data Processing and Training Details}

\paragraph{Data Processing}
Volumes were resampled to an in-plane resolution of $1.27\times1.27$~mm, with a slice 
spacing of $2.5$~mm. Each volume was cropped to a fixed size of $256\times256\times192$ using 
the center of the bounding box of the high-dose PTV as the volume center. Voxels 
outside the patient external contour, including couch components and immobilization
devices, were removed using masking. Dose and prescription values were linearly 
rescaled from $[0,80]$~Gy to $[-1,1]$; CT intensities were clipped to $[-1024,1900]$ 
HU and linearly rescaled to $[-1,1]$.

\paragraph{Dose Variational Autoencoder}

VAEs use downsampling factors $8\times8\times4$ and were trained patch-wise 
using Adam (lr $10^{-4}$, linear warm-up, $\beta=1.0$, batch size 8) on 4 RTX~3090 GPUs. 
Following the training process described in Section~\ref{sec:denoreg}, we train the dose VAE with latent 
corruption for 2800 epochs, then freeze the encoder and further fine-tune the decoder for 560 epochs. The interpolation noise scale is set to $\gamma=1.0$ to match the learned latent distribution. 

\paragraph{3D Diffusion Model} The diffusion model was trained on full volumes for 
140,000 steps on a single RTX~3090 with Adam (lr $10^{-4}$, linear warm-up, 
batch size 8) and the noise schedule of \citet{karrasElucidatingDesignSpace2022}. Inference leverages a deterministic ODE, 
Heun-EDM sampling and 20 denoising steps. Predictions were clamped to 
$[0, 1.06\times \text{prescription}]$. 

\subsubsection{Experiment Design}

\paragraph{Ablation Study}
We perform an incremental ablation study to assess the contribution of each component. To the baseline, we gradually introduce conditioning disentanglement (DisCond), latent denoising regularization (DenoReg) and multiscale classifier-free guidance (mCFG). Each variant was trained with the same schedule, budget and random seed. Evaluation was conducted under the same protocol as described in Section~\ref{sec:eval}, on the test set. By construction, classifier-free guidance operates only at inference; the corresponding ablation compares sampling with per-modality scales against sampling with a single scale applied to the combined conditioning set.

\paragraph{Evaluation on the AAPM GDP-HMM Dataset} For this experiment, we keep our autoencoder unchanged and retrain only the denoising network under the same budget and hyperparameter configuration. The latent space thus remains the one learned on our in-house dataset. The challenge's model is evaluated as released. Our model operates at a different resolution; we resample our predictions onto the challenge's grid, placing the cost of the interpolation on our model. 

\paragraph{Inference-Time Controllability Study} 

Inference-time controllability is characterized by multiple linear regression analysis between guidance scales and normalized dose quality metrics. We sampled a subset of 20 patients from the test set cohort at random, and generated 144 predictions per patient with guidance scales $(s_{\mathrm{CT}}, s_{\mathrm{PTV}}, s_{\mathrm{OAR}})$ drawn from a grid spanning $[1.0, 2.2]$ for the CT scale and $[1.0, 4.0]$ for the PTV and OAR scales. All other hyperparameters were fixed. DVH metrics were standardized with zero mean and unit variance before being regressed against each individual scale to allow for direct magnitude comparison. Monotonicity of the response was assessed by varying one scale at a time over its specific range, aggregating and averaging the resulting DVH metrics across the different scale combinations. 

\subsection{End-to-End Plan Generation}

\subsubsection{Direct Aperture Optimization Implementation Details} All plans were generated with two coplanar full arcs ($-179^{\circ}\rightarrow179^{\circ}$ clockwise and back counter-clockwise) with static collimator angles of $-30^{\circ}$ and $+30^{\circ}$, a control point every $2^{\circ}$ (179 per arc), dose-influence matrices computed every $8^{\circ}$ on a $5$~mm bixel grid, and a single isocenter placed at the centroid of the target volumes. Plans were computed for three different LINAC models: an Elekta Versa~HD with a flattened 6~MV beam, a Varian Halcyon with a 6~MV flattening-filter-free beam, and a Varian TrueBeam with a high-definition MLC and a flattened 6~MV beam. Each patient was planned using the configuration of the machine that delivered the treatment, so that each automated plan and its associated clinical plan could be compared directly. Additionally, the clinically approved, delivered plans were recalculated using the same collapsed-cone dose engine.

\subsubsection{Blinded Clinical Review and Quantitative Comparison}

As a proof of concept, the complete automated planning framework was assessed in a blinded reader study. Thirty patients were drawn at random from the test set, restricted to those treated from 2020 onwards to ground the study in contemporary clinical practice. Each patient contributed a single plan, either the delivered clinical plan or the automated plan, assigned at random (15 per arm); plans were never presented side by side, so that reviewers could not determine provenance by comparison. Four reviewers from three countries -- three senior radiation oncologists and one senior medical physicist -- each assessed a random subset of 20 of the 30 plans, blinded to plan provenance and to one another's assessments. Reviewers were shown the calculated plans along with DVH metrics and compliance with clinical constraints, using a conventional treatment planning system interface developed in-house. They rated each plan as acceptable as is, acceptable after minor editing, or not acceptable.

Independently of the reader study, the 30 plan pairs were analyzed quantitatively. For each patient, the automated plan and the delivered clinical plan were compared on the basis of standard DVH metrics, computed from the exported DICOM RT Plan objects after dose recalculation using a collapsed-cone-based dose engine. Total MU were compared as a surrogate for plan complexity and delivery efficiency. Statistical significance was tested with two-sided Wilcoxon signed-rank tests.

\section{Results}\label{sec:results}

We evaluate LaDose in three stages. First, we quantify the reconstruction fidelity of the dose variational autoencoder, including the effect of latent denoising regularization and its transfer to the AAPM GDP-HMM external cohort. We then assess the generative model's dose prediction accuracy on the in-house test set and the external cohort, together with ablation, generalization, and controllability analyses. Finally, we examine the clinical relevance of our approach by assessing the quality and acceptability of the plans generated by our end-to-end planning pipeline integrated with LaDose as a predictor. 

\subsection{Dose Latent Representation}

\paragraph{Reconstruction Capacity} Because the decoder bounds the performance of any denoiser on its latent space, we first characterize the learned representation. Table~\ref{tab:vae} reports the agreement between the delivered dose and its reconstruction through the encoder and decoder of Section~\ref{sec:latent} under the evaluation protocol described in Section~\ref{sec:eval}. At a compression rate of 128, the autoencoder reproduces the delivered dose with a median MAE of 0.135~Gy and a median DVH score of 0.160~Gy, well below the prediction error of the best-performing denoiser (Table~\ref{tab:sota_comparison}). This confirms that the representation does not limit the downstream predictive accuracy on our in-house dataset. 

\begin{figure*}
	\centering
	\includegraphics[width=0.9\textwidth]{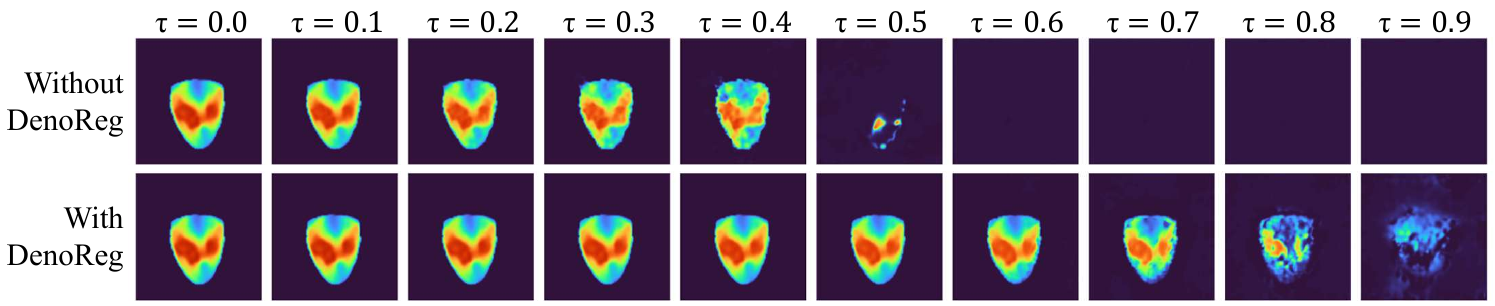}
	\caption{Decodability of the perturbed latents, with or without denoising regularization (DenoReg). Each dose embedding is interpolated with Gaussian noise following Eq.~\eqref{eq:latent_corruption} and decoded; columns show increasing corruption levels. Without regularization, reconstruction degrades beyond $\tau=0.2$; it remains stable until $\tau=0.6$ with denoising regularization.
  }
	\label{fig:denoreg}
\end{figure*}

\paragraph{Impact of Latent Denoising Regularization}\label{sec:denoreg_eval} 
Latent denoising regularization trades voxel-wise reconstruction fidelity for decodability of perturbed latents. We examine its impact on the reconstruction quality before examining its effect on generation. Indeed, training the decoder on strongly corrupted latents degraded voxel-wise reconstruction from 0.135~Gy to 0.217~Gy, while fine-tuning on clean samples recovered part of this loss, yielding a median MAE of $0.160$~Gy ($p=2.5\times10^{-14}$). The complete procedure remains worse voxel-wise than the baseline, but significantly improves reconstruction in PTVs and OARs, as illustrated by the removal of the systematic under-coverage of the high-dose PTV (from $-0.139$~Gy to $+0.039$~Gy in $D_{99}$, $p=0.02$; no detectable bias for the complete protocol only). The improvement is also reflected in the conformity index, which remains biased but falls by a factor of 5 (from $-0.0139$ to $+0.0026$, $p=0.02$). These improvements, which are more significant clinically than voxel-wise reconstruction, are consistent with the design rationale described in Section~\ref{sec:denoreg}. The gain is made even clearer when looking at generative performance (see Section~\ref{sec:ablation}), as the denoising pre-training component is the largest contributor to prediction quality. The trade of voxel-wise reconstruction fidelity for a more structured and learnable latent distribution is consistent with what \citet{yang2026latent} report for natural images. Figure~\ref{fig:denoreg} gives an illustration of the effect of latent denoising regularization on the reconstruction of dose distributions from a latent under different levels of corruption. 

\begin{figure*}
	\centering
	\includegraphics[width=0.9\textwidth]{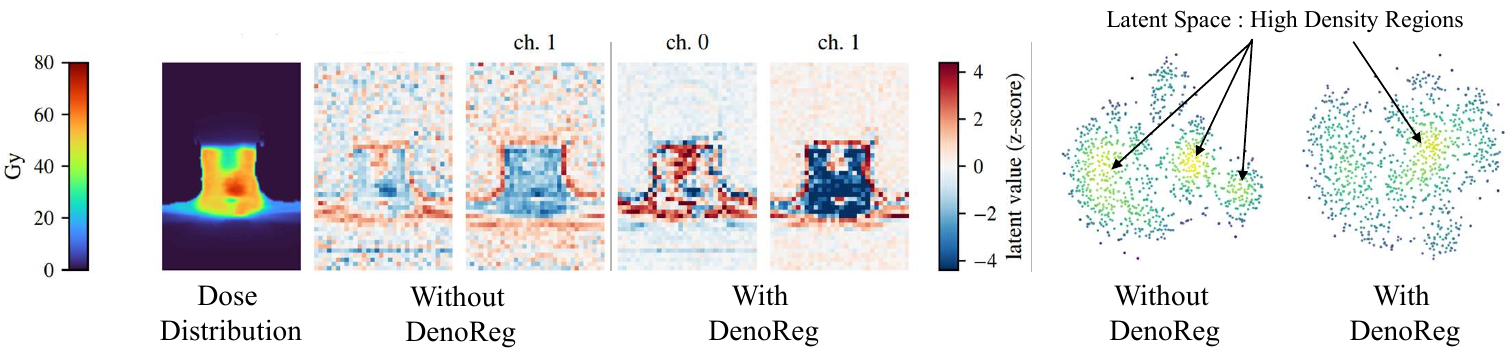}
	\caption{
    Effect of latent denoising regularization (DenoReg) on the learned representations and latent space structure. \textbf{Left:} coronal view of a 3D dose distribution and the associated embeddings, with and without DenoReg. The regularized representation is smoother and its structures are more contrasted. \textbf{Right:} 2D t-SNE embedding of the training set in the latent space, computed after reduction to 150 principal components (perplexity 30) and colored by kernel density estimate, high density regions appearing in yellow.
  }
	\label{fig:latentstruct}
\end{figure*}

Figure~\ref{fig:latentstruct} provides a qualitative assessment of the impact of denoising regularization on the dose embeddings and the structure of the latent space. The representation is visibly smoother and more structured, and the 2D projection of the complete training set in the latent space concentrates into fewer high density regions. \citet{maetok} empirically show that latent distributions with fewer modes are easier to learn. While our comparison remains qualitative, it supports the notion that denoising regularization encourages a more organized and learnable latent space.

\begin{table*}[t]
\centering
\caption{Fidelity of the dose autoencoder. Reconstruction of the delivered dose through the encoder and decoder of Section~\ref{sec:latent}, under the protocol of Section~\ref{sec:eval}. The two in-house columns differ only in how the decoder was trained: without latent denoising, versus with latent corruption followed by fine-tuning on clean samples. The last column applies that configuration, without retraining, to the public GDP-HMM cohort. Dose and DVH scores are absolute errors; $\Delta$ values are signed deviations from the delivered dose. Median $(Q_1,\,Q_3)$ across patients; values in bold mark significant superiority, $^{\dagger}$ marks a deviation significantly different from zero (two-sided Wilcoxon signed-rank, $\alpha=0.05$).}
\label{tab:vae}
\begin{threeparttable}
\footnotesize
\setlength{\tabcolsep}{4pt}
\renewcommand{\arraystretch}{1.1}
\begin{tabular}{@{}l cc c@{}}
\toprule
& \multicolumn{2}{c}{In-house test set}
& \multicolumn{1}{c}{GDP-HMM} \\
\cmidrule(lr){2-3} \cmidrule(lr){4-4}
Metric & Baseline & DenoReg (Ours)
& DenoReg (Ours) \\
\midrule
MAE (Gy) $\downarrow$
  & \mq{\mathbf{0.135}}{0.120}{0.153}
  & \mq{0.160}{0.147}{0.181}
  & \mq{0.312}{0.27}{0.36} \\
DVH score (Gy) $\downarrow$
  & \mq{0.160}{0.139}{0.185}
  & \mq{\mathbf{0.127}}{0.107}{0.180}
  & \mq{0.250}{0.19}{0.32} \\
\addlinespace
$\Delta$HI (pp)
  & \mq[\dagger]{-0.249}{-0.503}{-0.077}
  & \mq[\dagger]{-0.292}{-0.625}{-0.072}
  & \mq[\dagger]{+0.220}{-0.092}{+0.652} \\
$\Delta$nCI
  & \mq[\dagger]{-0.0139}{-0.0223}{-0.0091}
  & \mq[\dagger]{+0.0026}{-0.0005}{+0.0069}
  & \mq[\dagger]{-0.0242}{-0.0506}{-0.0141} \\
$\Delta D_{99}^{\mathrm{PTV}}$ (Gy)
  & \mq[\dagger]{-0.139}{-0.182}{-0.055}
  & \mq{\mathbf{+0.039}}{-0.065}{+0.154}
  & \mq[\dagger]{-0.381}{-0.736}{+0.061} \\
\bottomrule
\end{tabular}
\begin{tablenotes}[flushleft]\footnotesize
\item DenoReg: latent denoising regularization; HI: homogeneity index; nCI: Nakamura conformity index; pp: percentage points. 
\end{tablenotes}
\end{threeparttable}
\end{table*}

\subsection{Dose Prediction Performance}

\paragraph{Overall Accuracy on the In-House Test Set}

\begin{figure*}
	\centering
	\includegraphics[width=\textwidth]{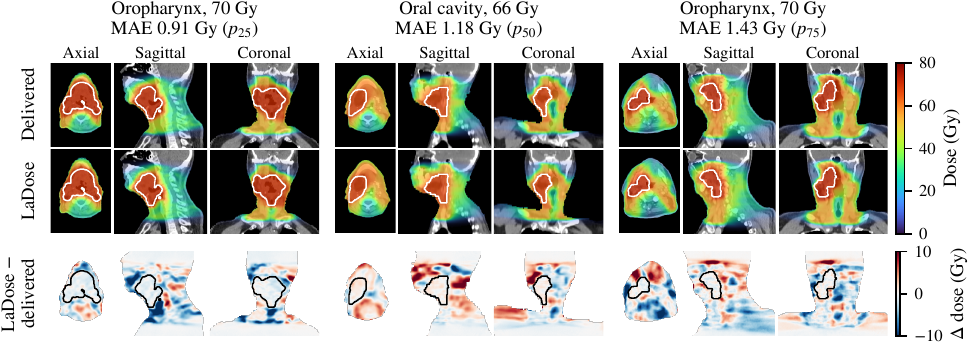}
	\caption{Qualitative comparison on three test patients. Each patient is shown on axial, sagittal and coronal slices through the center of the high-dose PTV. Rows, from top: the delivered clinical dose, the dose predicted by LaDose, and their difference (LaDose $-$ delivered). Cases are the patients at the $25$th, $50$th and $75$th percentile of the LaDose MAE distribution over the test set, so that they span the range of performance summarized in Table~\ref{tab:sota_comparison} rather than illustrating the best results. Prescriptions to the high-dose PTV are 70, 66 and 70~Gy. The high-dose PTV is outlined in white.}
	\label{fig:examples}
\end{figure*}

Table~\ref{tab:sota_comparison} summarizes the performance of LaDose against the five baselines on the held-out test set; Fig.~\ref{fig:examples} illustrates examples of dose distributions predicted by LaDose. Our model achieved the lowest voxel-wise error among published baselines, with a median MAE of $1.18$~Gy ($p\leq1.3\times10^{-8}$); the retrained GDP-HMM challenge winner, MedNeXt, matched it at $1.16$~Gy, a difference that is not significant ($p=1.00$). On the DVH score, LaDose performed significantly better than C3D, DCNN, MD-Dose ($p\leq1.6\times10^{-8}$) and MedNeXt ($p=1.6\times10^{-3}$), while being statistically indistinguishable from DoseDiff. No baseline was competitive on both metrics simultaneously: MedNeXt matched LaDose on MAE but not on the DVH score, and DoseDiff the opposite. 

In addition to the OpenKBP aggregate metrics, we further assess performance using the conventional radiotherapy metrics described in Section~\ref{sec:eval}. DCNN systematically under-covered the high-dose target by a median of $2.43$~Gy, an offset that the MAE column did not uncover. C3D, DCNN, DoseDiff and MedNeXt all predicted more conformal doses than were delivered ($\Delta$nCI $-0.090$, $-0.109$, $-0.040$, $-0.059$ respectively), while MD-Dose predicted a less conformal one ($\Delta$nCI $0.025$); all these deviations are significant. With regard to OAR sparing, both diffusion baselines systematically over-dosed the brainstem and spinal cord: DoseDiff by $2.11$~Gy and $1.97$~Gy, and MD-Dose by $2.63$~Gy and $1.70$~Gy. In contrast, C3D systematically under-dosed the spinal cord and parotids by $-2.13$~Gy and $-0.67$~Gy respectively. Because it can set an unachievable target, under-prediction can be a cause of optimization failure in a dose-mimicking pipeline. 

LaDose was the only model whose predictions did not differ significantly from the delivered plans across the seven metrics. By contrast, C3D and MedNeXt were biased on three, DCNN on four, DoseDiff on six and MD-Dose on five. The largest shift produced by our model was $+0.46$~Gy for spinal cord $D_{0.1\mathrm{cc}}$ (95\% CI $-0.45$ to $+1.42$), followed by $-0.48$~Gy for parotid $D_{\mathrm{mean}}$ ($-1.14$ to $+0.13$). Across all metrics, the confidence intervals exclude a systematic offset larger than $1.5$~Gy from the clinical plan; this indicates that LaDose allows for an accurate reproduction of the clinical dose across a heterogeneous cohort. 

\begin{table*}[t]
\centering
\caption{Evaluation of dose prediction performance on the test set. Dose and DVH scores are absolute errors; all other reported metrics are signed distances from the delivered clinical dose, with $\Delta=\text{predicted}-\text{clinical}$. The \emph{Clinical plan} column gives the absolute value of the delivered plan for reference. We report median $(Q_1,\,Q_3)$ across patients. Testing follows Section~\ref{sec:eval}: $^{*}$ marks a significant difference from LaDose,
$^{\dagger}$ a significant bias with respect to the delivered plan.}
\label{tab:sota_comparison}
\begin{threeparttable}
\footnotesize
\setlength{\tabcolsep}{1.5pt}
\renewcommand{\arraystretch}{1.1}
\begin{tabular}{@{}l c cccccc@{}}
\toprule
& Clinical & \multicolumn{6}{c}{Predicted dose} \\
\cmidrule(lr){2-2} \cmidrule(lr){3-8}
Metric &  & DCNN & C3D & DoseDiff & MD-Dose & MedNeXt & LaDose \\
& & \scriptsize Reg., 2D & \scriptsize Reg., 3D & \scriptsize Diff., 2D
  & \scriptsize Diff., 2D & \scriptsize Reg., 3D & \scriptsize Lat.\ diff., 3D \\
\midrule
\multicolumn{8}{@{}l}{\textit{Volumetric accuracy}}\\
MAE (Gy) & \nodata
  & \mq[*]{1.40}{1.07}{1.61} & \mq[*]{1.97}{1.73}{2.35}
  & \mq[*]{1.50}{1.22}{1.74} & \mq[*]{2.18}{1.81}{2.88}
  & \mq{1.16}{0.95}{1.48}
  & \mq{1.18}{0.90}{1.44} \\
DVH score (Gy) & \nodata
  & \mq[*]{1.81}{1.55}{2.08} & \mq[*]{1.64}{1.36}{2.11}
  & \mq{1.15}{0.91}{1.57} & \mq[*]{1.83}{1.36}{2.84}
  & \mq[*]{1.36}{0.97}{1.88}
  & \mq{1.19}{0.91}{1.57} \\
\addlinespace
\multicolumn{8}{@{}l}{\textit{Target coverage and dose shaping}}\\
$\Delta$HI (pp) & \mq{7.67}{5.99}{9.29}
  & \mq[*\dagger]{+2.14}{-0.36}{+3.65} & \mq[*]{-0.96}{-2.35}{+1.30}
  & \mq[\dagger]{-1.54}{-2.32}{-0.32} & \mq{+0.22}{-1.20}{+1.83}
  & \mq[*\dagger]{-1.43}{-3.07}{+0.29}
  & \mq{-0.09}{-1.29}{+1.24} \\
$\Delta$nCI & \mq{1.297}{1.216}{1.420}
  & \mq[*\dagger]{-0.109}{-0.181}{-0.030} & \mq[*\dagger]{-0.090}{-0.147}{-0.014}
  & \mq[\dagger]{-0.040}{-0.082}{+0.001} & \mq[*\dagger]{+0.052}{-0.015}{+0.293}
  & \mq[*\dagger]{-0.059}{-0.121}{+0.018}
  & \mq{-0.023}{-0.069}{+0.020} \\
$\Delta D_{99}^{\mathrm{PTV}}$ (Gy) & \mq{64.32}{60.79}{66.29}
  & \mq[*\dagger]{-2.43}{-3.21}{-0.70} & \mq{+0.08}{-0.59}{+1.27}
  & \mq[\dagger]{+0.32}{-0.35}{+1.13} & \mq{-0.15}{-1.00}{+0.76}
  & \mq[*\dagger]{+0.89}{-0.50}{+1.81}
  & \mq{+0.07}{-0.72}{+1.12} \\
\addlinespace
\multicolumn{8}{@{}l}{\textit{OAR sparing}}\\
$\Delta D_{0.1\mathrm{cc}}^{\mathrm{BS}}$ (Gy) & \mq{20.46}{12.50}{27.09}
  & \mq{-0.21}{-3.44}{+2.52} & \mq{-0.84}{-4.93}{+1.26}
  & \mq[\dagger]{+2.11}{-0.34}{+4.46} & \mq[*\dagger]{+2.63}{-0.60}{+8.02}
  & \mq{+0.87}{-2.14}{+3.23}
  & \mq{+0.02}{-1.86}{+2.34} \\
$\Delta D_{0.1\mathrm{cc}}^{\mathrm{SC}}$ (Gy) & \mq{29.82}{25.62}{33.44}
  & \mq{-0.58}{-3.24}{+2.50} & \mq[\dagger]{-2.13}{-5.21}{+0.36}
  & \mq[\dagger]{+1.97}{-0.63}{+4.41} & \mq[\dagger]{+1.70}{-0.24}{+4.61}
  & \mq{-0.11}{-3.28}{+2.45}
  & \mq{+0.67}{-2.19}{+3.03} \\
$\Delta D_{\mathrm{mean}}^{\mathrm{Par}}$ (Gy) & \mq{24.04}{19.92}{30.12}
  & \mq{-0.07}{-2.26}{+2.14} & \mq[\dagger]{-0.67}{-3.86}{+1.03}
  & \mq{-0.19}{-1.54}{+1.34} & \mq[\dagger]{+0.91}{-0.78}{+4.11}
  & \mq{-0.17}{-1.59}{+2.54}
  & \mq{-0.48}{-2.03}{+1.24} \\
$\Delta D_{0.1\mathrm{cc}}^{\mathrm{Mand}}$ (Gy) & \mq{63.21}{53.77}{69.29}
  & \mq[\dagger]{-0.49}{-2.13}{+0.60} & \mq{-0.27}{-1.51}{+0.44}
  & \mq[\dagger]{-0.68}{-1.21}{+0.26} & \mq[\dagger]{+0.51}{-0.67}{+2.38}
  & \mq{-0.13}{-1.09}{+1.02}
  & \mq{+0.03}{-1.03}{+1.03} \\
\addlinespace
\midrule
Metrics with a significant bias & --
  & 4 / 7 & 3 / 7 & 6 / 7 & 5 / 7 & 3 / 7 & \textbf{0 / 7} \\
\bottomrule
\end{tabular}
\begin{tablenotes}[flushleft]\footnotesize
\item BS: brainstem; SC: spinal cord; Par: parotid glands;
Mand: mandible; PTV: highest-dose target volume; pp: percentage points.
\end{tablenotes}
\end{threeparttable}
\end{table*}

\paragraph{Evaluation on the AAPM GDP-HMM Cohort} 

By retraining our denoiser on the AAPM GDP-HMM cohort, we aim to place the model on a public benchmark rather than assert superiority over the challenge winner. Indeed, our model is deliberately not conditioned on delivery parameters (see Section~\ref{sec:cond}) which does not enable relevant direct comparison.

The behavior observed on our in-house dataset is reproduced (Table~\ref{tab:aapm}); the challenge winner is more accurate voxel-wise (median MAE $1.23$ against $1.57$~Gy), while LaDose is more accurate on dosimetric endpoints (median DVH score $0.88$ against $1.03$~Gy). The difference is concentrated in tissue that carries no dosimetric constraint, in which the dose distribution is highly dependent on the geometry of the delivery. 

LaDose predicts a target dose $1.13$ percentage points less homogeneous than the reference dose and under-covers the high-dose target by $0.80$~Gy, while the challenge model is biased in the opposite direction, and significantly on brainstem near-maximum dose. Part of our model's deviation can be explained by the decoder's reconstruction bias: decoding the delivered dose through the autoencoder shifts $D_{99}$ by $-0.38$~Gy and HI by $+0.22$ percentage points, which is consistent with the observed bias in the predicted dose and the difference in planning style between the two cohorts.

\begin{table}[t]
\centering
\caption{Comparison on the public AAPM GDP-HMM head-and-neck validation cohort,
against the challenge-winning entry run from the authors' released checkpoint. Dose and DVH scores are absolute errors; all
other metrics are signed distances from the reference dose,
$\Delta=\text{predicted}-\text{reference}$. Median $(Q_1,\,Q_3)$ across patients. The last column
gives the two-sided Wilcoxon signed-rank $p$ between the two methods after Holm correction;
$^{\dagger}$ marks a significant bias with respect to the reference dose.}
\label{tab:aapm}
\begin{threeparttable}
\footnotesize
\setlength{\tabcolsep}{3pt}
\renewcommand{\arraystretch}{1.1}
\begin{tabular}{@{}l c cc c@{}}
\toprule
& Reference & \multicolumn{2}{c}{Predicted dose} & \\
\cmidrule(lr){2-2} \cmidrule(lr){3-4}
Metric & plan & MedNeXt & LaDose & $p_{\mathrm{Holm}}$ \\
& & \scriptsize Reg., 3D & \scriptsize Lat.\ diff., 3D & \\
\midrule
\multicolumn{5}{@{}l}{\textit{Volumetric accuracy}}\\
MAE (Gy) & \nodata
  & \mq{\mathbf{1.23}}{0.91}{1.45} & \mq{1.57}{1.20}{1.94} & $1.5\!\times\!10^{-12}$ \\
DVH score (Gy) & \nodata
  & \mq{1.03}{0.78}{1.32} & \mq{\mathbf{0.88}}{0.70}{1.03} & $0.011$ \\
\addlinespace
\multicolumn{5}{@{}l}{\textit{Target coverage and dose shaping}}\\
$\Delta$HI (pp) & \mq{4.27}{3.86}{5.12}
  & \mq[\dagger]{-0.66}{-1.58}{+0.30} & \mq[\dagger]{+1.13}{+0.51}{+1.86} & $4.1\!\times\!10^{-11}$ \\
$\Delta$nCI & \mq{1.461}{1.316}{1.637}
  & \mq{-0.018}{-0.129}{+0.017} & \mq[\dagger]{-0.034}{-0.082}{+0.004} & $1.00$ \\
$\Delta D_{99}^{\mathrm{PTV}}$ (Gy) & \mq{68.62}{64.78}{69.55}
  & \mq{+0.28}{-0.35}{+1.15} & \mq[\dagger]{-0.80}{-1.48}{-0.30} & $6.2\!\times\!10^{-10}$ \\
\addlinespace
\multicolumn{5}{@{}l}{\textit{OAR sparing}}\\
$\Delta D_{0.1\mathrm{cc}}^{\mathrm{BS}}$ (Gy) & \mq{22.22}{11.60}{32.63}
  & \mq[\dagger]{+0.93}{-0.30}{+2.29} & \mq{+0.51}{-1.66}{+2.07} & $0.065$ \\
$\Delta D_{0.1\mathrm{cc}}^{\mathrm{SC}}$ (Gy) & \mq{26.76}{24.87}{29.32}
  & \mq{-0.60}{-1.90}{+0.85} & \mq{-0.44}{-2.13}{+0.89} & $1.00$ \\
$\Delta D_{\mathrm{mean}}^{\mathrm{Par}}$ (Gy) & \mq{40.21}{31.08}{45.44}
  & \mq{+0.14}{-1.32}{+0.75} & \mq{+0.19}{-0.64}{+1.33} & $1.00$ \\
$\Delta D_{0.1\mathrm{cc}}^{\mathrm{Mand}}$ (Gy) & \mq{66.74}{60.44}{71.58}
  & \mq{-0.11}{-0.82}{+0.57} & \mq{-0.15}{-0.68}{+0.65} & $1.00$ \\
\addlinespace
\midrule
Metrics with a significant bias & -- & 2 / 7 & 3 / 7 & \\
\bottomrule
\end{tabular}
\begin{tablenotes}[flushleft]\footnotesize
\item BS: brainstem; SC: spinal cord; Par: parotid glands;
Mand: mandible; PTV: highest-dose target volume; pp: percentage points.
\end{tablenotes}
\end{threeparttable}
\end{table}

\paragraph{Ablation of Model Components}\label{sec:ablation}

Table~\ref{tab:ablation} reports the contribution of each component, evaluated under the protocol described in Section~\ref{sec:eval}. Disentangled conditioning lowered the median MAE from $1.34$ to $1.28$~Gy and improved $51$ of the $76$ test
patients ($p=0.010$ before correction), but the difference did not survive correction ($p_{\mathrm{Holm}}=0.09$). Its contribution is therefore consistent in direction on voxel-wise accuracy, and we maintain the component on the design grounds given in Section~\ref{sec:methods}.
Latent denoising regularization significantly improved prediction performance, lowering the median MAE to $1.18$~Gy ($p_{\mathrm{Holm}}=5.2\times10^{-10}$) and the DVH score from $1.34$ to $1.19$~Gy ($p_{\mathrm{Holm}}=1.6\times10^{-5}$). Regarding dosimetric endpoints, target homogeneity and parotid mean dose improved ($p_{\mathrm{Holm}}=0.049$ and $0.042$). Both configurations trained without it predicted a target dose consistently less homogeneous than delivered ($\Delta$HI $+1.05$ and $+0.88$ percentage points, $p_{\mathrm{Holm}}<10^{-3}$), the only endpoint on which either was biased; with it the deviation falls to $-0.09$ percentage points ($p_{\mathrm{Holm}}=1.00$), with no significant bias. Denoising pre-training acts on the quality of the dose representation rather than on the denoising network. The result supports the claim of the present work: the fidelity of the prediction is highly dependent on the learned latent space. 

Because multiscale guidance acts only at inference, the last two rows sample the checkpoint of the preceding row with per-modality scales rather than a single scale. The effect sizes quoted below are medians, across the test patients, of the within-patient difference between the two configurations, which is the paired quantity the Wilcoxon test is computed on; they therefore do not in general equal the difference between the corresponding cells of Table~\ref{tab:ablation}, each of which is a cohort median taken independently. Raising $s_{\mathrm{PTV}}$ from $1.4$ to $1.8$ increased target coverage ($D_{99}$ $+0.31$~Gy, $p_{\mathrm{Holm}}=3.5\times10^{-12}$), improved homogeneity (HI $-0.24$~pp) and conformity (nCI $-0.011$), while raising parotid $D_{\mathrm{mean}}$ by $0.15$~Gy; MAE, DVH score, brainstem and spinal cord maximum doses were left unchanged ($p_{\mathrm{Holm}}=1.00$). In contrast, raising $s_{\mathrm{OAR}}$ to $4.0$ reversed the trade: all four OAR dosimetric endpoints fell significantly (spinal cord $D_{0.1\mathrm{cc}}$ $-0.82$~Gy, mandible $-0.59$~Gy, brainstem $-0.63$~Gy, parotids $-0.48$~Gy, all $p_{\mathrm{Holm}}<10^{-8}$), concurrently and significantly lowering target coverage ($D_{99}$ $-0.53$~Gy).

\begin{table*}[t]
\centering
\caption{Incremental ablation of the model components, on the same test set and under the protocol of Section~\ref{sec:eval}. Each cell is a cohort median over the test patients: the absolute error for MAE and the DVH score, and the signed deviation of the prediction from the delivered plan for the $\Delta$ endpoints. Clinical reference values for the deviation endpoints are those of Table~\ref{tab:sota_comparison}. $^{*}$ marks a significant difference from the preceding configuration, $^{\dagger}$ a significant bias with respect to the delivered plan, $^{\ddagger}$ a significant difference from the single-scale reference (third row). The last two rows share the checkpoint of the third and differ only in the inference-time guidance scales; they characterize controllability rather than accuracy, and are therefore not marked with $^{*}$.}
\label{tab:ablation}
\begin{threeparttable}
\footnotesize
\setlength{\tabcolsep}{3pt}
\begin{tabular}{@{}ccc c cc ccc cccc@{}}
\toprule
\multicolumn{3}{c}{Components} & Guidance & \multicolumn{2}{c}{Error (Gy)}
& \multicolumn{3}{c}{Target} & \multicolumn{4}{c}{OAR sparing (Gy)} \\
\cmidrule(lr){1-3} \cmidrule(lr){4-4} \cmidrule(lr){5-6} \cmidrule(lr){7-9} \cmidrule(lr){10-13}
DisCond & DenoReg & mCFG & $s_{\mathrm{CT}}/s_{\mathrm{PTV}}/s_{\mathrm{OAR}}$
& MAE $\downarrow$ & DVH $\downarrow$
& $\Delta$HI & $\Delta$nCI & $\Delta D_{99}$
& $\Delta$BS & $\Delta$SC & $\Delta$Par & $\Delta$Mand \\
\midrule
  & & & $1.4/1.4/1.4$ & $1.34$ & $1.39$
  & $+1.05^{\dagger}$ & $0.000$ & $-0.11$
  & $-0.18$ & $+1.34$ & $-0.12$ & $+0.33$ \\
\accYes & & & $1.4/1.4/1.4$ & $1.28$ & $1.34$
  & $+0.88^{\dagger}$ & $-0.004$ & $+0.12$
  & $-0.37$ & $+1.25$ & $-0.56$ & $+0.27$ \\
\accYes & \accYes & & $1.4/1.4/1.4$ & $\mathbf{1.18}^{*}$ & $\mathbf{1.19}^{*}$
  & $\mathbf{-0.09}^{*}$ & $-0.023$ & $+0.07$
  & $+0.02$ & $+0.67$ & $-0.48^{*}$ & $+0.03$ \\
\midrule
\accYes & \accYes & \accYes & $1.4/1.8/1.4$ & $1.20$ & $1.13$
  & $-0.22^{\ddagger}$ & $-0.029^{\dagger\ddagger}$ & $+0.33^{\ddagger}$
  & $-0.05$ & $+0.54$ & $-0.32^{\ddagger}$ & $+0.11$ \\
\accYes & \accYes & \accYes & $1.4/1.4/4.0$ & $1.22$ & $1.27$
  & $+0.38$ & $-0.040^{\dagger}$ & $-0.62^{\ddagger}$
  & $-0.41^{\ddagger}$ & $-0.28^{\ddagger}$ & $-0.59^{\dagger\ddagger}$ & $-0.56^{\dagger\ddagger}$ \\
\bottomrule
\end{tabular}
\begin{tablenotes}[flushleft]\footnotesize
\item DisCond: disentangled conditioning; DenoReg: latent denoising regularization;
mCFG: multiscale classifier-free guidance. HI in percentage points, nCI
dimensionless, all other endpoints in Gy. Structure abbreviations as in
Table~\ref{tab:sota_comparison}.
\item \accYes\ marks a component present in that configuration; an empty cell marks it absent.
\end{tablenotes}
\end{threeparttable}
\end{table*}

\paragraph{Generalization Across Tumor Sites and Prescriptions}

\begin{figure*}
  \centering
  \includegraphics[width=\textwidth]{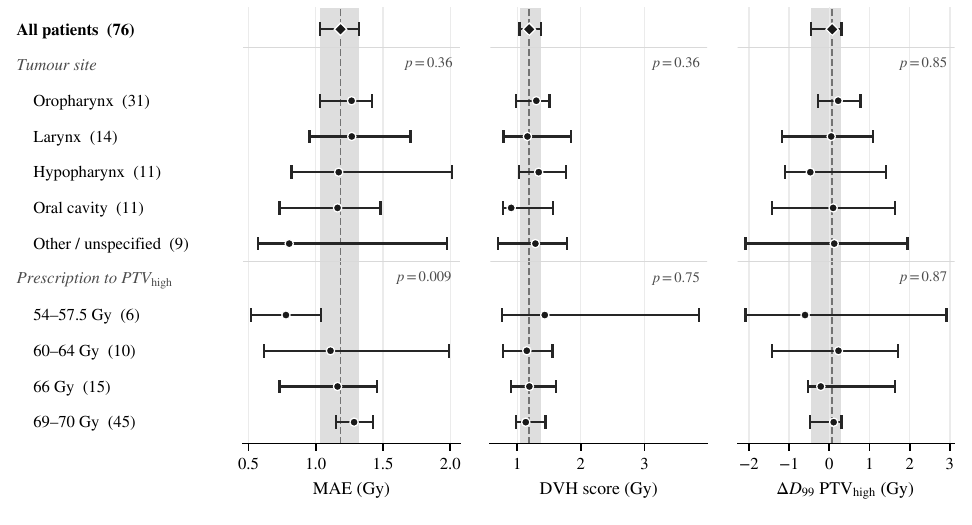}
  \caption{Prediction performance on the test set ($n=76$) stratified by primary tumor site and prescription. Points are group medians, bars their $95\%$ distribution-free confidence intervals (achieved coverage $96$--$99\%$, as such an interval can only fall on observed values). In each panel the dashed line and the gray band are the whole-cohort median and 95\% confidence interval. The italic $p$ is the Kruskal--Wallis test of heterogeneity across the subgroups for that metric.}
  \label{fig:generalization}
\end{figure*}

Most published dose prediction models are developed on cohorts limited to a single tumor site and prescription protocol (see Section~\ref{sec:related}), while LaDose is designed to generalize across the head-and-neck region. In addition to aggregate metrics which could mask subgroup-specific failures, we stratify the test cohort by primary tumor site and nominal prescription dose and report associated evaluation metrics. Results are reported in Fig.~\ref{fig:generalization}. The model's performance was consistent across the five primary tumor sites (Kruskal--Wallis $p \geq 0.36$), and every subgroup median lay within $0.4$~Gy of the whole-cohort median on both error metrics. Target coverage remained unbiased, with the largest deviation being $-0.47$~Gy in the hypopharynx group ($n=11$, 95\% CI $-1.11$ to $+1.42$). Across prescription groups, the DVH metrics behaved similarly, while the voxel-wise error increased with the primary target prescription value from a median MAE of $0.78$~Gy at $54$--$57.5$~Gy to $1.28$~Gy at $69$--$70$~Gy ($p = 0.009$). When expressed as a fraction of the prescription, the error is $1.4$--$1.8\%$ in every group and the difference is no longer detectable ($p = 0.16$), indicating that the increase in absolute error is a reflection of the scale of the quantity being predicted rather than a loss of accuracy. The remaining metrics, reported in Tables~\ref{tab:gen-site} and~\ref{tab:gen-rx}, showed no heterogeneity either ($p \geq 0.075$); only $D_{99}$ is plotted, being the only metric whose subgroup intervals are small enough to be informative ($95\%$ half-width $0.5$--$1.5$~Gy in the four named sites, against up to $7$~Gy for the OAR endpoints). With $6$ to $45$ patients per stratum and site and prescription level inherently dependent, the analysis is descriptive. The results support the claim that no tumor site and no prescription protocol behaves as a systematic failure mode of the model.

\paragraph{Inference-Time Controllability via Multiscale Guidance}

Control over the predicted dose at inference is what enables a single trained model to adapt to the planning preferences of users from different institutions (see Section~\ref{sec:related}). We therefore characterize inference-time controllability using multiple linear regression analysis between guidance scales and normalized dose quality metrics. Significance is reported after Holm correction across the metric--scale pairs. The PTV scale exhibited the strongest influence on the prediction, sharpening the gradient at the tumor boundary (nCI: $\beta = -0.72$, 95\% confidence interval $[-0.82, -0.62]$, $p_{\mathrm{Holm}} < 0.001$), with the cohort mean nCI decreasing monotonically from $1.33$ to $1.27$ across $s_{\mathrm{PTV}} \in [1, 4]$. As expected, the same scale acted in the opposite direction on normal tissue, increasing parotid $D_{\mathrm{mean}}$ ($\beta = 0.87$, $[0.84, 0.90]$), brainstem $D_{\mathrm{max}}$ ($\beta = 0.83$, $[0.81, 0.85]$) and spinal cord $D_{\mathrm{max}}$ ($\beta = 0.70$, $[0.67, 0.73]$). The OAR scale acted in the opposite direction on all three (spinal cord $D_{\mathrm{max}}$: $\beta = -0.76$, $[-0.79, -0.74]$; brainstem $D_{\mathrm{max}}$: $\beta = -0.64$, $[-0.66, -0.62]$; parotid $D_{\mathrm{mean}}$: $\beta = -0.51$, $[-0.55, -0.48]$), at the cost of target coverage (PTV$_{\text{high}}$ $D_{99}$: $\beta = -0.63$, $[-0.71, -0.54]$). The CT scale was the only one whose effect on OAR dose was negligible ($|\beta| \leq 0.10$ for all three organs); it acted instead on agreement with the clinical plan (dose score: $\beta = -0.69$, $[-0.74, -0.63]$). The PTV and OAR channels therefore act as opposing controls on the trade-off between tumor coverage and OAR sparing, while the anatomical channel is the only one that does not displace dose between them.

\begin{figure*}
  \centering
	\includegraphics[width=\textwidth]{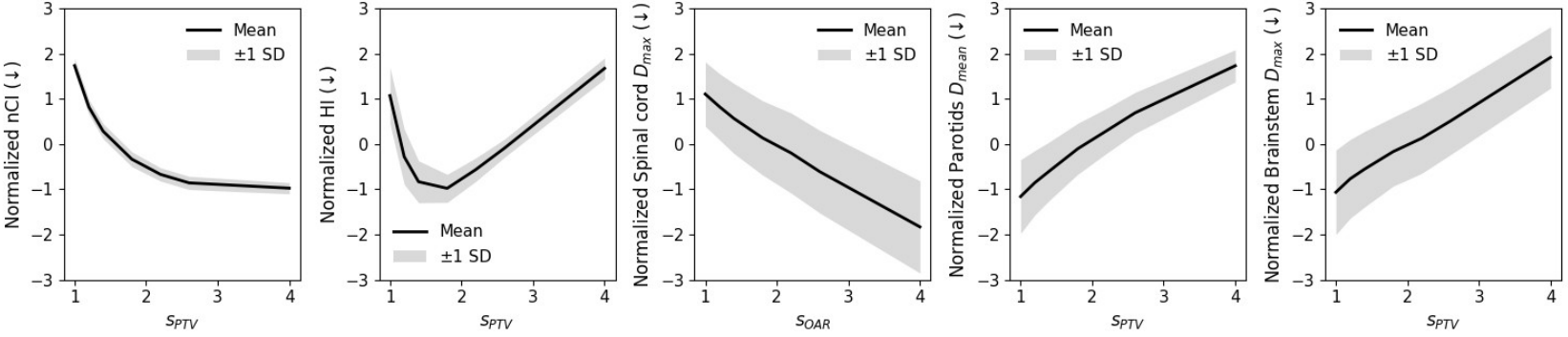}
	\caption{Effect of guidance scales on dose quality metrics. Increasing the PTV guidance
scale $s_{\mathrm{PTV}}$ improves the conformity of target coverage (nCI) at the expense of brainstem and parotid sparing. Increasing the OAR guidance scale $s_{\mathrm{OAR}}$ encourages spinal cord sparing. Solid lines show mean normalized values and shaded regions indicate $\pm1$ standard deviation.}
  \label{fig:response_curves}
\end{figure*}

The linear model explained most of the variation in the dose quality metrics, with $R^2 = 0.91$ for the dose score, $0.66$ for the conformity index, and $0.99$, $0.98$ and $0.97$ for brainstem $D_{\mathrm{max}}$, spinal cord $D_{\mathrm{max}}$ and parotid $D_{\mathrm{mean}}$ respectively. Homogeneity is the one metric whose response is not monotonic: it improves up to $s_{\mathrm{PTV}} \approx 1.8$ and degrades beyond it, with cohort mean values spanning $7.87$--$8.52$ percentage points; we report HI as a response curve only. Fig.~\ref{fig:response_curves} shows the corresponding response curves, which illustrate how CFG enables the predictable modulation of clinical trade-offs at inference.

\subsection{End-to-End Automated Treatment Planning}

\paragraph{Quantitative Comparison with Clinical Plans}

A predicted dose is not a treatment plan; it is used as a voxel-wise objective that an optimizer converts into a deliverable set of machine parameters (see Section~\ref{sec:dao}). The quality of the plan depends on both the predictor and the optimizer's performances \citep{babierOpenKBPOptInternationalReproducible2022}. We therefore investigate clinical applicability of LaDose by integrating it into an end-to-end automated treatment planning pipeline and evaluating the resulting 30 plans against their delivered counterparts.

Our ATP pipeline produced the 30 plans from planning CT, prescription and contours with a median planning time of 15.2~min. Automated plans were significantly less modulated than their delivered counterparts, with a median within-patient reduction of 67~MU per fraction ($-9.6\%$ of the clinical median of 693~MU; mean 617 versus 729~MU) and a standard deviation of 77 versus 198~MU, consistent with the fixed two-arc geometry. The framework achieved superior OAR sparing (Table~\ref{tab:quant}), with median reductions of 4.1~Gy to the right parotid, 2.4~Gy to the left parotid, 2.4~Gy to the mandible, and 3.8~Gy to the spinal cord and 1.6~Gy to the brainstem in near-maximum dose. The overall maximum dose received by the patient was reduced by 2.2~Gy. Target coverage was maintained, with high-dose PTV $D_{95}$ only 0.57~Gy lower in the automated plans (95\% CI $-0.85$ to $-0.40$~Gy) and all plans satisfying the clinical constraint of $D_{95}>95\%$ of the prescription.

The automated plans were generated using the available contours. Inter-institutional protocol discrepancies result in heterogeneous contouring practices. The results summarized in Table~\ref{tab:quant} consider only the structures contoured in every patient. The results for structures which were not systematically contoured can be found in the appendix (Table~\ref{tab:quant-appendix}).

\begin{table*}[t]
\centering
\caption{Dosimetric and plan-complexity comparison of automated and delivered clinical
plans, for the structures contoured in essentially every patient. Values are
mean\,$\pm$\,SD across patients; $\Delta$ is the median within-patient difference
(automated $-$ clinical) with a 95\% percentile confidence interval. Doses in Gy.}
\label{tab:quant}
\begin{threeparttable}
\small
\setlength{\tabcolsep}{4pt}
\begin{tabular}{@{}llcccc@{}}
\toprule
Structure & Metric & Clinical & Automated & $\Delta$ (95\% CI) & $p_{\mathrm{Holm}}$ \\
\midrule
\multicolumn{6}{@{}l}{\textit{Organs at risk}}\\
\addlinespace
Left parotid & $D_{\mathrm{mean}}$ & $23.24\pm13.31$ & $20.16\pm12.93$ & $-2.43$ \cinterval{-4.36}{-1.11} & $5.3\times10^{-5}$ \\
Right parotid & $D_{\mathrm{mean}}$ & $24.03\pm11.60$ & $20.03\pm11.15$ & $-4.15$ \cinterval{-5.41}{-2.66} & $5.3\times10^{-5}$ \\
Oral cavity & $D_{\mathrm{mean}}$ & $36.28\pm16.41$ & $34.51\pm16.12$ & $-1.37$ \cinterval{-3.50}{+0.04} & $0.0482$ \\
Spinal cord & $D_{2}$ & $30.07\pm5.28$ & $25.75\pm5.74$ & $-3.79$ \cinterval{-5.66}{-2.03} & $6.2\times10^{-5}$ \\
Brainstem & $D_{2}$ & $15.00\pm10.95$ & $13.30\pm9.77$ & $-1.57$ \cinterval{-2.79}{-0.24} & $0.0482$ \\
Mandible & $D_{\mathrm{max}}$ & $60.87\pm17.67$ & $58.28\pm17.49$ & $-2.36$ \cinterval{-3.25}{-1.73} & $1.4\times10^{-4}$ \\
\addlinespace
\multicolumn{6}{@{}l}{\textit{Targets and global maximum}}\\
\addlinespace
PTV high dose & $D_{98}$ & $65.78\pm3.79$ & $65.43\pm3.86$ & $-0.37$ \cinterval{-0.81}{-0.07} & $0.0691$ \\
 & $D_{95}$ & $66.84\pm3.80$ & $66.23\pm3.82$ & $-0.57$ \cinterval{-0.85}{-0.40} & $3.5\times10^{-4}$ \\
 & $D_{2}$ & $71.48\pm4.07$ & $69.23\pm4.00$ & $-2.31$ \cinterval{-2.90}{-1.70} & $4.7\times10^{-8}$ \\
 & HI & $8.40\pm1.76$ & $5.60\pm1.51$ & $-2.31$ \cinterval{-3.70}{-1.36} & $2.0\times10^{-7}$ \\
PTV low dose & $D_{98}$ & $53.31\pm3.83$ & $52.67\pm3.82$ & $-0.75$ \cinterval{-1.18}{-0.45} & $0.0166$ \\
 & $D_{95}$ & $54.60\pm3.68$ & $53.97\pm3.62$ & $-0.74$ \cinterval{-0.98}{-0.54} & $0.00266$ \\
 & $D_{2}$ & $70.18\pm4.18$ & $68.48\pm4.06$ & $-1.67$ \cinterval{-2.15}{-1.46} & $1.6\times10^{-6}$ \\
 & HI & $30.88\pm10.22$ & $29.00\pm9.83$ & $-1.82$ \cinterval{-2.24}{-0.95} & $0.0011$ \\
Body & $D_{\mathrm{max}}$ & $73.50\pm4.16$ & $71.07\pm4.16$ & $-2.22$ \cinterval{-3.19}{-1.64} & $5.0\times10^{-7}$ \\
\addlinespace
\multicolumn{6}{@{}l}{\textit{Plan complexity}}\\
\addlinespace
Plan & MU / fraction & $729\pm198$ & $617\pm77$ & $-67$ \cinterval{-185}{-4} & $0.0166$ \\
\bottomrule
\end{tabular}
\end{threeparttable}
\end{table*}

\paragraph{Blinded Expert Acceptability Study}
Automatic plans were systematically rated as clinically acceptable, with 14 out of 15 requiring no editing (against 1 out of 15 for the delivered plans---the majority did not reach consensus; Table~\ref{tab:aggregate}). In contrast, four clinical plans were found to require re-planning by at least one reviewer, against none for the automated plans. Automated planning favored consensus among clinicians, with experts agreeing with one another on 13 of the 14 automated plans carrying two independent ratings or more, against 2 out of 14 clinical plans ($p=6.9\times10^{-5}$; Table~\ref{tab:aggregate}). This proof-of-concept study demonstrates that our end-to-end planning pipeline is able to produce plans that are clinically acceptable without editing according to state-of-the-art clinical practice. Further clinical evaluation is out of this project's scope, but is a promising future direction. 

\begin{table}[t]
\centering
\caption{Clinical acceptability of automatically generated versus manually generated, clinically delivered plans. A plan is counted as acceptable only if every reviewer assigned to it rated it acceptable as is. Agreement is computed on the plans carrying two or more independent ratings. Proportions with 95\% Wilson score intervals; $p$ from two-sided Fisher exact tests.}
\label{tab:aggregate}
\begin{threeparttable}
\small
\begin{tabular}{@{}lccc@{}}
\toprule
& Automated & Clinical & $p$ \\
\midrule
Acceptable as is, all reviewers
  & 14/15 \; 93\% \cinterval{70}{99} & 1/15 \; 7\% \cinterval{1}{30} & $2.9\times10^{-6}$ \\
Re-planning required, $\geq$1 reviewer
  & 0/15 \; 0\% \cinterval{0}{20} & 4/15 \; 27\% \cinterval{11}{52} & $0.10$ \\
\addlinespace
Complete reviewer agreement\tnote{a}
  & 13/14 \; 93\% \cinterval{69}{99} & 2/14 \; 14\% \cinterval{4}{40} & $6.9\times10^{-5}$ \\
\bottomrule
\end{tabular}
\begin{tablenotes}[flushleft]\footnotesize
\item[a] Restricted to the 14 plans per arm carrying $\geq2$ independent ratings; all assigned reviewers gave the same rating category, whichever category that was.
\end{tablenotes}
\end{threeparttable}
\end{table}

\section{Discussion}\label{sec:discussion}

\paragraph{Radiotherapy-Specific Latent Representation} In their concurrent work, \citet{wang_any2any_2026} also perform volumetric latent diffusion, but obtain their representation by transfer from an autoencoder pretrained on natural videos. The latter is kept frozen, while the diffusion transformer is adapted to the dose prediction task. The present work takes a different path and learns the representation on dose distributions directly. Two consequences follow. Our framework leverages a smaller denoising network (1.3~B for the Wan~2.1 diffusion transformer, against 67~M for ours), which makes training and design iterations more tractable. Because the representation is not required to encode the high fidelity details of natural scenes, it allows for a substantially higher compression rate (128 versus 16 for Any2Any).

\paragraph{Conditioning Signal Design Choices}\label{sec:cond}

Models supplied with beam configuration conditioning consistently improve on reconstruction metrics \citep{teng_beam-wise_2024, cao_beam_2025, xiongGeneralizableDosePrediction2026}. We deliberately omit this signal from our framework as the delivery geometry is an output of the treatment planning procedure rather than an input naturally available at prediction time. Nonetheless, the present work is compatible with beam-conditioned models, and the proposed framework could be extended to include beam geometry in the future.

\paragraph{Clinical Relevance of the Fully Automated Planning Pipeline}
We do not claim that the quality of the plans produced by the end-to-end pipeline is directly linked to the performance of our dose prediction model. We consider the complete pipeline as a unit, and claim that its different components work together to produce plans deemed clinically acceptable by senior clinicians. The moderate correlation between dose prediction metrics and plan quality reported by \citet{babierOpenKBPOptInternationalReproducible2022} is why we report both rather than treat plan quality as a corollary of predictive accuracy. 

\paragraph{Limitations and Future Work}

Several limitations of the present work should be acknowledged. In our current implementation, guidance scales are an abstract quantity which may be difficult for clinicians to interpret. A slider which maps this scale to a concrete DVH objective, together with per-organ rather than a pooled OAR scale, is a promising direction for future work. We decide to remove the variability inherent to sampling with a diffusion model during inference to ensure reproducibility at the cost of diversity. Studying this stochasticity and its potential benefits for uncertainty-aware planning is a natural extension. Finally, the end-to-end pipeline was compared against clinical plans but not against plans generated by the same optimizer integrated with a different predictor, so the contribution of prediction fidelity to overall plan quality cannot be formally isolated.

\section{Conclusion}

In this work, we presented and evaluated LaDose, a fully volumetric, prescription-conditioned, controllable, latent-diffusion-based dose prediction model for head-and-neck VMAT radiotherapy. We demonstrated that learning a structured dose latent representation enabled tractable 3D diffusion on a single GPU, and that aligning the decoder with the denoising task rather than treating it as a preprocessing step significantly contributed to predictive accuracy (Table~\ref{tab:ablation}). Our multiscale guidance mechanism allowed for predictable modulation of the trade-off between target coverage and OAR sparing at inference, which addresses inter-clinician variability without requiring preferences to be specified before training (Fig.~\ref{fig:response_curves}). LaDose reached a median absolute error of 1.18~Gy (1.4--1.8\% of the prescription), and was the only model of six without a statistically significant bias in any evaluated dosimetric endpoint. The plans produced by our end-to-end treatment planning pipeline with a median time of 15.2 minutes were rated clinically acceptable without editing in 14 of 15 cases (Table~\ref{tab:aggregate}) in a blinded review by three senior radiation oncologists and one medical physicist across three countries.

\appendix

\section{Reimplementation of Baseline Methods}\label{app:baselines}

Baselines were retrained from scratch on our in-house dataset with the authors' released implementation, and evaluated under the protocol of Section~\ref{sec:eval}. Three main adaptations were implemented: the field of view and voxel grid were set to the $256\times256\times192$ crop at $1.27\times1.27\times2.5$~mm used throughout this work; prescription conditioning was introduced wherever the released implementation assumed a uniform prescription protocol. The structure channels were remapped to the five contours available in our cohort. Every method is scored against the same reference dose and the same structures. In each case, the checkpoint minimizing the validation dose score was retained. DCNN required no further adaptation, while other model-specific changes are detailed below.

\paragraph{C3D} is the two-stage cascade of 3D U-Nets that won the OpenKBP
challenge \citep{C3D, babierOpenKBPOpenaccessKnowledgebased2021a}. In order to train the model on our higher resolution grid, we trained C3D patch-wise, with a crop containing the PTVs (as in our model). Inference uses a sliding window with Gaussian weighting. 

\paragraph{DoseDiff} is a slice-wise conditional DDPM ($T=1000$, DDIM sampling with $S=8$; \citealp{dosediff}). We normalize the dose and PTV channels by 80~Gy rather than the 75~Gy in the paper, as the maximum delivered dose in our cohort is 77.4~Gy. We match and follow three elements of the released training schedule that the paper does not describe: a 1000-step learning-rate warm-up, gradient clipping, and a single $\times0.1$ learning-rate decay.

\paragraph{MD-Dose} is a slice-wise conditional DDPM ($T=1000$) whose Mamba-UNet denoiser is conditioned on a five-channel structure image (CT, three PTV masks, OAR label map) \citep{fuMDDoseDiffusionModel2025}. Four adaptations were made, in order to address a discrepancy between the released code and the paper's description or change a setting under which the model did not train. First, we normalized the dose and PTV channels by a fixed 80~Gy rather than the released per-patient min--max, which is not suitable to a multi-prescription setting. Second, the released implementation samples the initial latent from a min--max rescaled latent of mean $0.5$
and standard deviation $0.1$, whereas training injects $\varepsilon\sim\mathcal{N}(0,\mathbf{I})$; we matched inference to training sampling, which significantly improved the validation MAE. Third, the reported initial learning rate of $10^{-2}$ diverged on our data, so we adopted the $10^{-4}$ of the released code. Fourth, our training split yields over $7000$ optimizer steps per epoch against
roughly $1250$ for the authors' cohort; we match their training budget in steps rather than epochs to make retraining tractable.

\paragraph{MedNeXt} is the winning entry of the AAPM 2025 GDP-HMM challenge \citep{gdpwinner}. The challenge introduces two additional delivery geometry input channels, which were zeroed out in our reimplementation. We note that the system that won the challenge is a voxel-wise equal-weight ensemble of this model with the organizers' baseline network; we reproduce the single MedNeXt model only.

\section{Quantitative Evaluation of the End-to-End Planning Pipeline}\label{app:quant-appendix}

Due to the variability in contouring practices, not every organ at risk is delineated in every patient, which affects the sample size for statistical testing. In Table~\ref{tab:quant-appendix}, we report the quantitative comparison of the automated and delivered clinical plans for the organs at risk and target volumes that were not systematically delineated. Structures delineated in at least 24 out of 30 patients were tested with two-sided exact Wilcoxon signed-rank tests, whereas those delineated in fewer patients are described in Table~\ref{tab:quant-appendix} without testing, as the sample size does not support it. 

\begin{table*}[t]
\centering
\caption{Dosimetric comparison of automated and delivered clinical plans for the organs
at risk and target volumes that were not delineated in every patient. Columns are as in
Table~\ref{tab:quant}. Structures delineated in at least 24 of the 30 patients were tested
with two-sided exact Wilcoxon signed-rank tests and the Benjamini--Hochberg procedure
applied across that family; structures delineated in fewer patients are reported
descriptively only, as the available sample does not support a meaningful test. Doses in Gy.}
\label{tab:quant-appendix}
\begin{threeparttable}
\small
\setlength{\tabcolsep}{4pt}
\begin{tabular}{@{}llccccc@{}}
\toprule
Structure & Metric & $n$ & Clinical & Automated & $\Delta$ (95\% CI) & $q_{\mathrm{BH}}$ \\
\midrule
\multicolumn{7}{@{}l}{\textit{Delineated in $\ge$24/30 patients: tested}}\\
\addlinespace
Larynx & $D_{\mathrm{mean}}$ & 25 & $47.06\pm18.35$ & $44.91\pm19.05$ & $-2.10$ \cinterval{-3.62}{-1.00} & $0.0507$ \\
 & $D_{5}$ & 25 & $64.03\pm13.79$ & $60.87\pm15.51$ & $-2.32$ \cinterval{-3.59}{-1.77} & $7.2\times10^{-7}$ \\
Esophagus & $D_{\mathrm{mean}}$ & 27 & $10.50\pm7.92$ & $9.48\pm6.62$ & $-0.89$ \cinterval{-1.59}{-0.02} & $0.0301$ \\
 & $D_{5}$ & 27 & $31.92\pm17.25$ & $30.62\pm17.15$ & $-1.39$ \cinterval{-2.51}{+0.08} & $0.114$ \\
Left submandibular gland & $D_{\mathrm{mean}}$ & 28 & $49.00\pm21.35$ & $45.19\pm21.13$ & $-3.57$ \cinterval{-4.95}{-2.70} & $3.2\times10^{-4}$ \\
Right submandibular gland & $D_{\mathrm{mean}}$ & 24 & $54.53\pm16.58$ & $51.80\pm15.61$ & $-2.90$ \cinterval{-4.29}{-2.10} & $4.9\times10^{-4}$ \\
Left brachial plexus & $D_{\mathrm{max}}$ & 24 & $54.23\pm18.29$ & $51.57\pm19.59$ & $-2.02$ \cinterval{-2.73}{-1.62} & $2.2\times10^{-4}$ \\
Right brachial plexus & $D_{\mathrm{max}}$ & 24 & $54.92\pm16.59$ & $52.80\pm16.80$ & $-1.86$ \cinterval{-3.19}{-0.55} & $3.2\times10^{-4}$ \\
Left cochlea & $D_{\mathrm{mean}}$ & 28 & $7.57\pm9.53$ & $5.87\pm6.83$ & $-0.09$ \cinterval{-1.34}{+0.29} & $0.178$ \\
 & $D_{5}$ & 28 & $10.60\pm13.31$ & $8.30\pm10.62$ & $-0.10$ \cinterval{-2.24}{+0.29} & $0.137$ \\
Right cochlea & $D_{\mathrm{mean}}$ & 28 & $6.73\pm8.26$ & $5.02\pm5.89$ & $-0.09$ \cinterval{-0.93}{+0.28} & $0.114$ \\
 & $D_{5}$ & 28 & $10.27\pm14.02$ & $7.52\pm10.09$ & $-0.09$ \cinterval{-1.85}{+0.32} & $0.119$ \\
\addlinespace
\multicolumn{7}{@{}l}{\textit{Delineated in $<$24/30 patients: not tested}}\\
\addlinespace
Trachea & $D_{\mathrm{mean}}$ & 21 & $16.48\pm11.09$ & $14.59\pm10.56$ & $-1.46$ \cinterval{-2.25}{-0.12} & -- \\
 & $D_{5}$ & 21 & $39.30\pm18.93$ & $36.68\pm18.68$ & $-1.06$ \cinterval{-3.11}{+0.16} & -- \\
Hypophysis & $D_{\mathrm{mean}}$ & 19 & $2.71\pm2.26$ & $2.51\pm1.28$ & $+0.14$ \cinterval{-0.38}{+0.42} & -- \\
Pharyngeal constrictors & $D_{\mathrm{mean}}$ & 14 & $45.30\pm14.09$ & $43.05\pm14.15$ & $-2.10$ \cinterval{-3.60}{-1.39} & -- \\
PTV intermediate dose & $D_{95}$ & 4 & $53.66\pm0.84$ & $53.11\pm0.44$ & $-0.43$ \cinterval{-2.00}{+0.64} & -- \\
 & $D_{98}$ & 4 & $52.87\pm0.96$ & $52.59\pm0.46$ & $-0.20$ \cinterval{-1.96}{+1.22} & -- \\
 & $D_{50}$ & 4 & $62.43\pm4.00$ & $60.23\pm4.65$ & $-2.40$ \cinterval{-2.72}{-1.26} & -- \\
 & $D_{2}$ & 4 & $70.58\pm2.26$ & $69.19\pm2.50$ & $-1.40$ \cinterval{-2.13}{-0.65} & -- \\
\bottomrule
\end{tabular}
\begin{tablenotes}[flushleft]\footnotesize
\item Availability is a property of the clinical structure set and is therefore identical
for the two plans of a pair; each row is a complete paired sample on the patients in which
the structure was delineated.
\end{tablenotes}
\end{threeparttable}
\end{table*}

\section{Generalization Across Tumor Sites and Prescription Protocols}\label{app:generalization}

Figure~\ref{fig:generalization} plots MAE and DVH score, along with target coverage, as these are the only endpoints whose subgroup intervals are narrow enough to be significant. Tables~\ref{tab:gen-site} and~\ref{tab:gen-rx} give the full endpoint set over the same subgroups. These tests are exploratory and we draw no definitive conclusions from them.
\begin{table*}[t]
\centering
\caption{Prediction performance on the test set stratified by primary tumor site.
Values are median $(Q_1,\,Q_3)$ across patients. Dose and DVH scores are absolute
errors; all other endpoints are signed deviations from the delivered clinical plan,
$\Delta=\text{predicted}-\text{clinical}$. Structure abbreviations as in
Table~\ref{tab:sota_comparison}.}
\label{tab:gen-site}
\footnotesize
\setlength{\tabcolsep}{3pt}
\renewcommand{\arraystretch}{1.1}
\begin{tabular}{@{}l cccccc r@{}}
\toprule
Endpoint & \shortstack{All\\{\scriptsize$(n=76)$}} & \shortstack{Oropharynx\\{\scriptsize$(n=31)$}} & \shortstack{Larynx\\{\scriptsize$(n=14)$}} & \shortstack{Hypopharynx\\{\scriptsize$(n=11)$}} & \shortstack{Oral cavity\\{\scriptsize$(n=11)$}} & \shortstack{Other/unspecified\\{\scriptsize$(n=9)$}} & KW $p$ \\
\midrule
MAE $\downarrow$ (Gy) & \mq{1.18}{0.90}{1.44} & \mq{1.27}{0.93}{1.43} & \mq{1.27}{0.99}{1.46} & \mq{1.17}{0.97}{1.55} & \mq{1.16}{0.77}{1.35} & \mq{0.80}{0.61}{1.04} & 0.360 \\
DVH $\downarrow$ (Gy) & \mq{1.19}{0.91}{1.57} & \mq{1.30}{0.93}{1.54} & \mq{1.16}{0.86}{1.81} & \mq{1.34}{1.06}{1.59} & \mq{0.91}{0.84}{1.09} & \mq{1.29}{0.97}{1.63} & 0.359 \\
$\Delta$HI (pts) & \mq{-0.09}{-1.29}{+1.24} & \mq{-0.20}{-2.41}{+0.52} & \mq{-0.02}{-1.20}{+1.77} & \mq{+0.53}{-0.50}{+1.14} & \mq{-0.12}{-1.60}{+1.09} & \mq{+0.78}{-0.40}{+1.98} & 0.417 \\
$\Delta$nCI & \mq{-0.023}{-0.069}{+0.020} & \mq{-0.023}{-0.057}{+0.016} & \mq{+0.013}{-0.117}{+0.077} & \mq{-0.062}{-0.098}{-0.019} & \mq{-0.004}{-0.049}{+0.014} & \mq{-0.042}{-0.111}{+0.043} & 0.585 \\
$\Delta D_{99}^{\mathrm{PTV}}$ (Gy) & \mq{+0.07}{-0.72}{+1.12} & \mq{+0.22}{-0.65}{+1.13} & \mq{+0.05}{-1.05}{+0.40} & \mq{-0.47}{-0.75}{+0.21} & \mq{+0.10}{-0.53}{+1.29} & \mq{+0.12}{-2.07}{+1.60} & 0.848 \\
$\Delta D_{0.1\mathrm{cc}}^{\mathrm{BS}}$ (Gy) & \mq{+0.02}{-1.86}{+2.34} & \mq{-0.49}{-1.61}{+2.93} & \mq{+0.06}{-0.84}{+1.16} & \mq{-0.01}{-5.24}{+1.66} & \mq{+0.31}{-2.24}{+1.99} & \mq{+1.34}{-1.76}{+2.60} & 0.953 \\
$\Delta D_{0.1\mathrm{cc}}^{\mathrm{SC}}$ (Gy) & \mq{+0.67}{-2.19}{+3.03} & \mq{+1.50}{-1.46}{+3.61} & \mq{+1.00}{-2.02}{+3.13} & \mq{+0.63}{-2.99}{+2.53} & \mq{-1.33}{-3.40}{+0.63} & \mq{+1.45}{-0.32}{+2.05} & 0.411 \\
$\Delta D_{\mathrm{mean}}^{\mathrm{Par}}$ (Gy) & \mq{-0.48}{-2.03}{+1.24} & \mq{-0.33}{-1.66}{+1.75} & \mq{-0.11}{-2.29}{+1.28} & \mq{-2.47}{-6.34}{+0.11} & \mq{-0.50}{-1.47}{+1.59} & \mq{-0.50}{-1.11}{+0.21} & 0.478 \\
$\Delta D_{0.1\mathrm{cc}}^{\mathrm{Mand}}$ (Gy) & \mq{+0.03}{-1.03}{+1.03} & \mq{+0.21}{-0.20}{+1.38} & \mq{-0.62}{-4.60}{+1.37} & \mq{-0.84}{-3.49}{+0.25} & \mq{+0.09}{-0.28}{+0.48} & \mq{-0.05}{-1.38}{+1.17} & 0.139 \\
\bottomrule
\end{tabular}

\end{table*}

\begin{table*}[t]
\centering
\caption{Prediction performance on the test set stratified by the dose prescribed to
the high-dose PTV. Columns and conventions as in Table~\ref{tab:gen-site}.}
\label{tab:gen-rx}
\footnotesize
\setlength{\tabcolsep}{3pt}
\renewcommand{\arraystretch}{1.1}
\begin{tabular}{@{}l ccccc r@{}}
\toprule
Endpoint & \shortstack{All\\{\scriptsize$(n=76)$}} & \shortstack{54--57.5 Gy\\{\scriptsize$(n=6)$}} & \shortstack{60--64 Gy\\{\scriptsize$(n=10)$}} & \shortstack{66 Gy\\{\scriptsize$(n=15)$}} & \shortstack{69--70 Gy\\{\scriptsize$(n=45)$}} & KW $p$ \\
\midrule
MAE $\downarrow$ (Gy) & \mq{1.18}{0.90}{1.44} & \mq{0.78}{0.59}{0.85} & \mq{1.11}{0.76}{1.67} & \mq{1.16}{0.77}{1.36} & \mq{1.28}{1.03}{1.46} & 0.009 \\
DVH $\downarrow$ (Gy) & \mq{1.19}{0.91}{1.57} & \mq{1.43}{1.10}{1.71} & \mq{1.15}{0.91}{1.30} & \mq{1.19}{0.93}{1.59} & \mq{1.13}{0.91}{1.57} & 0.751 \\
$\Delta$HI (pts) & \mq{-0.09}{-1.29}{+1.24} & \mq{+1.80}{-1.17}{+2.93} & \mq{+0.19}{-1.15}{+3.39} & \mq{-0.12}{-2.33}{+1.69} & \mq{-0.10}{-1.21}{+0.76} & 0.544 \\
$\Delta$nCI & \mq{-0.023}{-0.069}{+0.020} & \mq{-0.032}{-0.064}{-0.013} & \mq{+0.060}{-0.014}{+0.100} & \mq{-0.023}{-0.050}{+0.017} & \mq{-0.033}{-0.084}{+0.006} & 0.075 \\
$\Delta D_{99}^{\mathrm{PTV}}$ (Gy) & \mq{+0.07}{-0.72}{+1.12} & \mq{-0.60}{-1.61}{+1.12} & \mq{+0.23}{-0.86}{+1.12} & \mq{-0.21}{-0.50}{+1.63} & \mq{+0.11}{-0.65}{+0.62} & 0.870 \\
$\Delta D_{0.1\mathrm{cc}}^{\mathrm{BS}}$ (Gy) & \mq{+0.02}{-1.86}{+2.34} & \mq{-0.37}{-1.24}{+2.04} & \mq{+0.30}{-1.41}{+1.40} & \mq{+0.64}{-1.73}{+3.42} & \mq{-0.23}{-3.79}{+2.39} & 0.915 \\
$\Delta D_{0.1\mathrm{cc}}^{\mathrm{SC}}$ (Gy) & \mq{+0.67}{-2.19}{+3.03} & \mq{-0.65}{-6.93}{+1.28} & \mq{+0.38}{-2.57}{+2.98} & \mq{+0.42}{-2.38}{+2.88} & \mq{+0.94}{-1.41}{+3.01} & 0.836 \\
$\Delta D_{\mathrm{mean}}^{\mathrm{Par}}$ (Gy) & \mq{-0.48}{-2.03}{+1.24} & \mq{+0.87}{-0.61}{+2.39} & \mq{-0.80}{-1.74}{+0.18} & \mq{-0.72}{-2.26}{+0.13} & \mq{-0.17}{-2.42}{+1.71} & 0.548 \\
$\Delta D_{0.1\mathrm{cc}}^{\mathrm{Mand}}$ (Gy) & \mq{+0.03}{-1.03}{+1.03} & \mq{-0.79}{-1.32}{-0.15} & \mq{+0.74}{-0.10}{+1.13} & \mq{+0.30}{-0.14}{+0.95} & \mq{-0.02}{-1.00}{+0.84} & 0.464 \\
\bottomrule
\end{tabular}

\end{table*}

\printcredits

\section*{Ethics approval and consent to participate}
This retrospective study was conducted in accordance with the Declaration of Helsinki. Data were anonymized by each center before transfer and provided to Therapanacea under data sharing agreements in compliance with the General Data Protection Regulation. Patients were informed of the possible reuse of their data for research purposes and did not object. The use of the public AAPM GDP-HMM dataset follows its terms of use.

\section*{Declaration of competing interest}
The authors declare the following financial interests/personal relationships which may be considered as potential competing interests: This work was funded by Therapanacea and the Association Nationale de la Recherche et de la Technologie (ANRT) through a CIFRE doctoral contract. Jeanne Boyer-Chammard reports a relationship with Therapanacea that includes: employment. Nikos Paragios reports a relationship with Therapanacea that includes: employment and equity or stocks. Remi Vauclin reports a relationship with Therapanacea that includes: employment. Nikos Komodakis reports a relationship with Therapanacea that includes: consulting or advisory. Jeanne Boyer-Chammard has patent pending to Therapanacea, Institut Gustave Roussy and Centre Léon Bérard. If there are other authors, they declare that they have no known competing financial interests or personal relationships that could have appeared to influence the work reported in this paper.

\section*{Declaration of generative AI and AI-assisted technologies in the manuscript preparation process}
During the preparation of this work the authors used ChatGPT (OpenAI) and Claude (Anthropic) in order to assist with drafting and improving the language and readability of the manuscript. After using these tools, the authors reviewed and edited the content as needed and take full responsibility for the content of the published article.

\section*{Data availability}
The authors do not have permission to share data. 

\section*{Acknowledgments}
This work is supported by Therapanacea and the ANRT through a CIFRE doctoral contract (Convention CIFRE no.~2024/0915). Therapanacea was involved in the study design, data analysis,
writing of the manuscript and the decision to submit it for publication.

\bibliographystyle{cas-model2-names}

\bibliography{cas-refs}

\end{document}